\documentclass[11pt]{article}

\usepackage[final]{acl}

\usepackage{times}
\usepackage{latexsym}

\usepackage[T1]{fontenc}

\usepackage[utf8]{inputenc}

\usepackage{microtype}

\usepackage{inconsolata}

\usepackage{graphicx}
\usepackage{booktabs}
\usepackage{multirow}
\usepackage{wrapfig}
\usepackage{subcaption}
\usepackage{listings}
\usepackage{longtable}
\usepackage{xcolor}
\usepackage{amsmath}
\usepackage{amssymb}

\usepackage{hyperref}
\usepackage{url}
\usepackage[utf8]{inputenc}
\usepackage{tabularray}
\usepackage{geometry}
\usepackage{comment}
\usepackage{todonotes}
\usepackage{amsmath}
\usepackage{amssymb}
\usepackage{color}
\usepackage[T1]{fontenc}
\usepackage{microtype}
\usepackage{inconsolata}
\usepackage{graphicx}
\usepackage{subcaption}
\usepackage{wrapfig}
\usepackage{siunitx}
\usepackage{listings}
\usepackage{caption}
\usepackage{pifont}
\usepackage{enumitem}

\usepackage{CJKutf8}
\lstdefinestyle{json}{
    language=json,
    basicstyle=\ttfamily\small,
    numbers=none,
    commentstyle=\color{gray},
    stringstyle=\color{blue},
    keywordstyle=\color{purple},
    identifierstyle=\color{black},
    showstringspaces=false,
    breaklines=true,
    frame=single,
    framerule=0.5pt,
    rulecolor=\color{gray!40},
    backgroundcolor=\color{gray!5},
}

\title{RedNote-Vibe: A Dataset for Capturing Temporal Dynamics of AI-Generated Text in Lifestyle Social Media}

\author{%
  Yudong Li\textsuperscript{\rm 1}
  \quad Yufei Sun \textsuperscript{\rm 2}
  \quad Peiru Yang\textsuperscript{\rm 1*}
  \quad Yuhan Yao\textsuperscript{\rm 2}
   \quad Wanyue Li\textsuperscript{\rm 3} 
   \quad \textbf{Jiajun Zou}\textsuperscript{\rm 1} \\
   \quad \textbf{Haoyang Yang}\textsuperscript{\rm 4} 
   \quad \textbf{Haotian Gan}\textsuperscript{\rm 1} 
    \quad \textbf{Linlin Shen}\textsuperscript{\rm 4} 
    \quad \textbf{Yongfeng Huang}\textsuperscript{\rm 1}  \\
  \textsuperscript{\rm 1}Tsinghua University
  \quad \textsuperscript{\rm 2}Beijing University of Posts and Telecommunications\\
  \textsuperscript{\rm 3} Hong Kong Metropolitan University
  \quad \textsuperscript{\rm 4}Shenzhen University \\
  \texttt{liyudong@tsinghua.edu.cn, ypr21@mails.tsinghua.edu.cn}
}

\begin{document}
\maketitle
\begin{abstract}
We introduce \textbf{RedNote-Vibe}, a dataset spanning five years (pre-LLM to July 2025) sourced from lifestyle platform RedNote (Xiaohongshu), capturing the temporal dynamics of content creation and is enriched with comprehensive engagement metrics. To address the detection challenge posed by RedNote-Vibe, we propose the \textbf{PsychoLinguistic AIGT Detection Framework (PLAD)}. Grounded in cognitive psychology, PLAD leverages deep psychological signatures for robust and interpretable detection. Our experiments demonstrate PLAD's superior performance and reveal insights into content dynamics: (1) human content continues to outperform AI in emotionally resonant domains; (2) AI content is more homogeneous and rarely produces breaking posts, however, this human-AI gap narrows for arousing higher-investment interactions; and (3) most interestingly, a small group of users who strategically utilize AI tools can achieve higher engagement outcomes. \url{https://github.com/ydli-ai/RedNote-Vibe}

\end{abstract}

\section{Introduction}

Large Language Models (LLMs) \citep{achiam2023gpt, google2025gemini2, guo2025deepseek} have revolutionized digital content creation, leading to a surge in AI-Generated Text (AIGT). Consequently, developing robust detection methods has become a critical research frontier, with a focus on classification \citep{gui2025aider, hu2023radar} and source attribution \citep{sun2025idiosyncrasies}. However, existing research primarily focus on knowledge-sharing and formal corpora (e.g., news, academic writing). There remains a lack of research on lifestyle-sharing social media platforms, where content is deeply intertwined with personal narratives and commercial influence.

In this work, we identify two critical and unaddressed challenges for AIGT research on social media. First, unlike formal text, where factual accuracy is prioritized, \textbf{social media rewards content that maximizes user engagement}, such as likes, comments, and shares \citep{chung2023really, cascio2024high}. On social media platforms that value sharing real-life experiences, LLMs can be prompted to generate sensationalized or controversial content to increase interactions, which undermines connections and trust within the community.
Second, \textbf{over time, AI and human content are interacting more frequently on social media platforms.} This longitudinal mixture of content sources presents challenges for detection methods, as linguistic patterns may shift over time. Existing datasets, typically static snapshots from informational sources (e.g., Reddit, Quora), fail to capture these temporal dynamics or the engagement patterns of commercialized social content.


To bridge this gap, we introduce RedNote-Vibe, the first dataset designed for studying AIGT in a dynamic, lifestyle-centric social media context. RedNote-Vibe is collected from RedNote (Xiaohongshu), a leading Chinese platform combining social networking with e-commerce. Our data spans a wide timeline (Oct 2020 - July 2025), covering the transition from the pre-LLM era to the post-LLM era. The dataset features: (1) rich metadata including engagement metrics (likes, comments, collections); (2) parallel AIGT variants generated by 17 diverse LLMs based on human seeds; and (3) an exploration set of real-world post-LLM content. It provides a natural testbed for observing the evolution of AIGT and its impact on user interaction.

Furthermore, facing the challenge that as LLMs evolve, surface-level linguistic artifacts (e.g., syntax patterns) will inevitably be neutralized. We propose the PsychoLinguistic AIGT Detection Framework (PLAD). Instead of relying solely on statistical patterns, PLAD leverages linguistic features grounded in psycholinguistic theories to construct interpretable detection models.

Our experiments demonstrate PLAD not only achieves superior performance compared to black-box baselines but also offers insights into how specific psychological signatures correlate with social media engagement. The main contributions of this paper are as follows:

\begin{figure}{}
    \centering
    \includegraphics[width=0.49\textwidth]{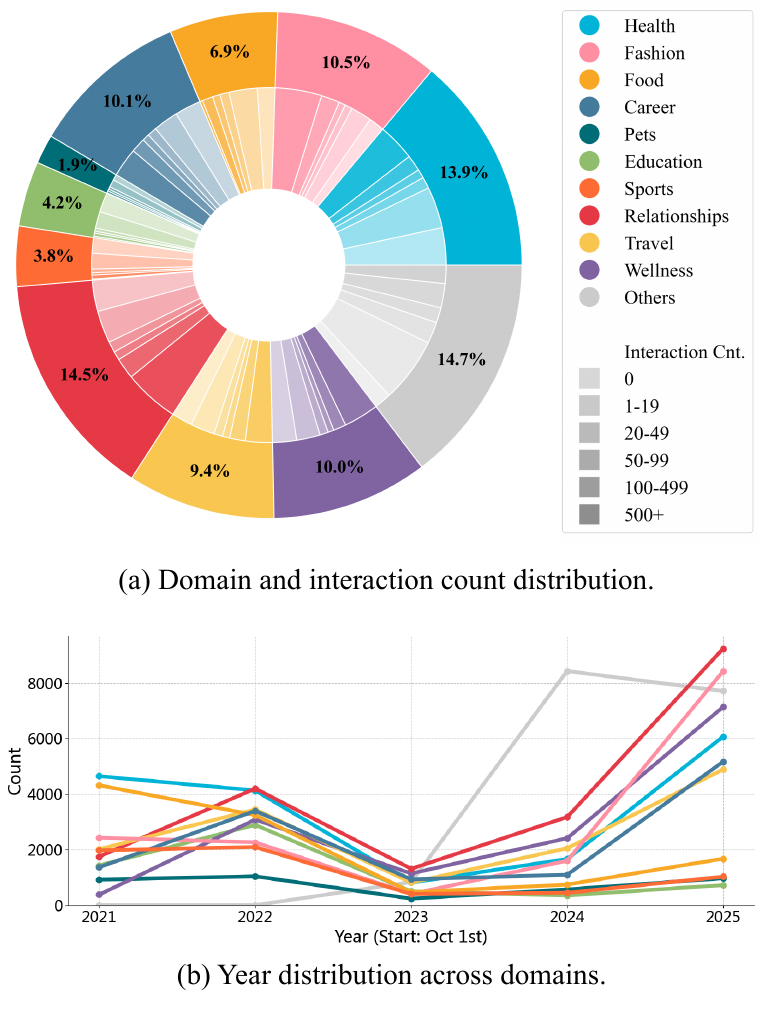}
    \caption{RedNote-Vibe dataset distribution.}
    \vspace{-0.3cm}

    \label{fig:dist}
\end{figure}

\begin{itemize}[itemsep=0pt, topsep=0pt, parsep=0pt]
    \item We introduce RedNote-Vibe, a large-scale Chinese social media AIGT dataset. To our knowledge, it is the first dataset from the RedNote platform.
    \item We propose PLAD, an interpretable framework leveraging psycholinguistic theories, offering robust detection and explaining the linguistic style of AIGT.
    \item We provide a comprehensive analysis of the dataset, uncovering temporal trends in AI adoption and revealing how AI-generated content differs from human authorship in driving social interactions.
 \end{itemize}

\section{RedNote Engagement Dataset}

\begin{figure*}[!t]
    \centering
    
    \includegraphics[width=\linewidth]{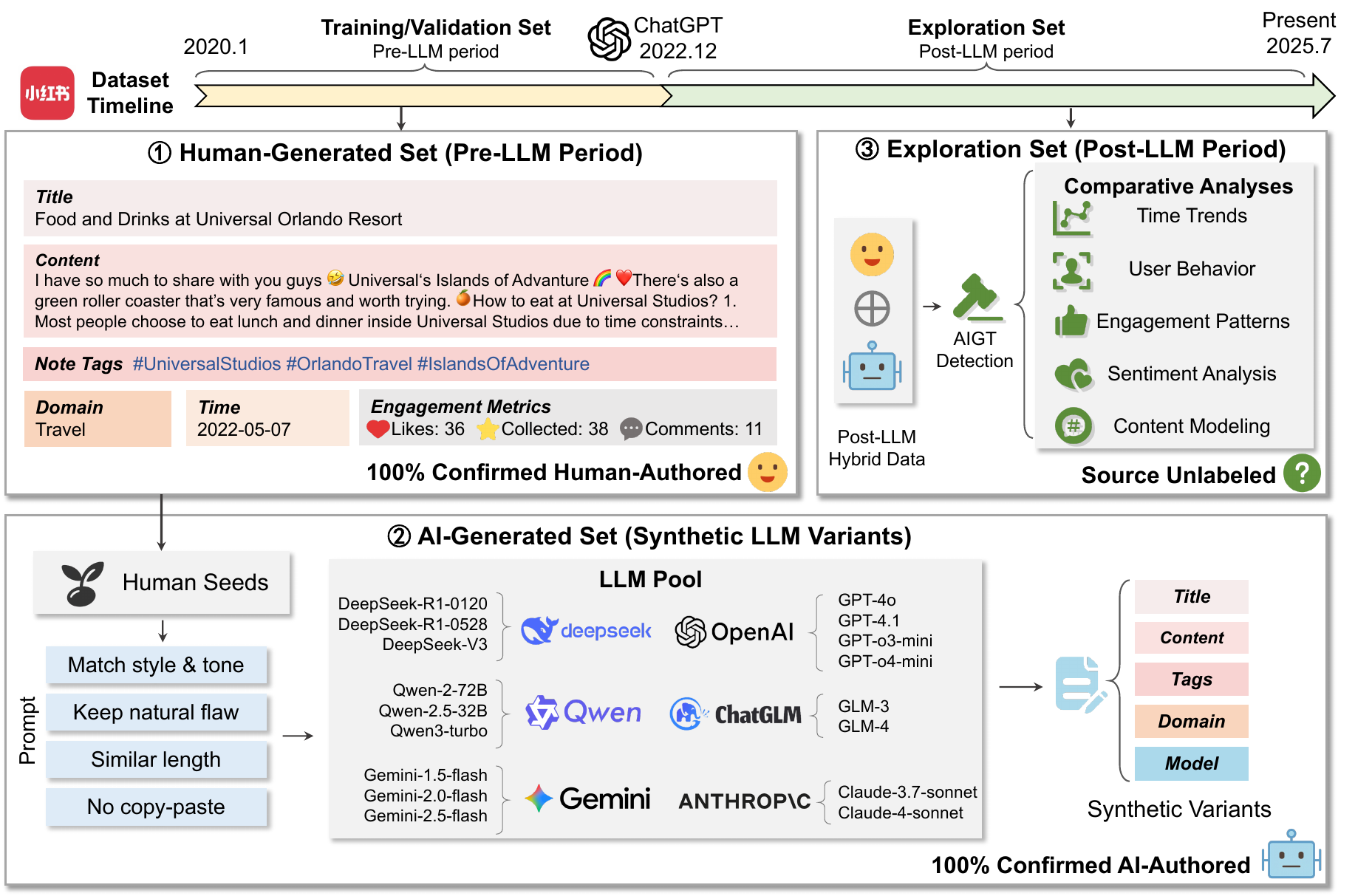}
    \caption{Overview of the RedNote-Vibe dataset construction. \textbf{(1) Human-Generated Set} is collected in the pre-LLM period before 2022.12, therefore labeled as human-authored. \textbf{(2) AI-Generated Set} is created by prompting a diverse LLM pool with the human seeds to produce synthetic variants. \textbf{(3) Exploration Set} contains post-LLM social posts, enabling AIGT detection and extensive analyses comparing AI and human content.}
    \label{fig:method}
    
\end{figure*}

RedNote (Xiaohongshu)\footnote{https://www.xiaohongshu.com} stands as one of the most influential Chinese social media platforms, serving over 300 million monthly active users. This platform emphasizes personal experiences and lifestyle sharing, which makes it particularly vulnerable to AIGC infiltration, as it undermines the authenticity of the content. Despite its influence, RedNote has remained largely unexplored in academic research due to the absence of a publicly available dataset. In this section, we present our RedNote-Vibe dataset, the first publicly available dataset from this platform. It also captures temporal dynamics and engagement patterns, specifically designed for research on AIGT detection and the impact of AIGC on social media.

\begin{table}[t]
\centering
\small

    \resizebox{\linewidth}{!}{
    \begin{tabular}{lcccccc}

    \toprule
    \textbf{Domain} & \textbf{\#} & \textbf{Length} & \textbf{\#Tags} & \textbf{Likes} & \textbf{Comm.} & \textbf{Colls.} \\
    \midrule
    Health        & 19.2k & 333.6 & 7.6 & 2814.3 & 97.1 & 2468.8 \\ 
    Fashion       & 17.7k & 191.2 & 10.9 & 1850.5 & 62.5 & 681.3 \\ 
    Food          & 10.5k  & 398.4 & 3.0 & 66.2   & 11.9 & 45.0 \\ 
    Career        & 11.7k & 406.9 & 10.0 & 2047.2 & 102.6 & 1445.9 \\ 
    Pets          & 3.9k  & 433.2 & 2.7 & 57.5   & 12.6 & 34.1 \\ 
    Education     & 5.9k  & 496.0 & 3.7 & 55.9   & 6.1  & 38.3 \\ 
    Sports        & 17.7k  & 528.8 & 3.2 & 60.1   & 9.8  & 36.4 \\ 
    Relation.     & 22.5k & 308.2 & 7.1 & 2663.6 & 275.8 & 593.9 \\ 
    Travel        & 14.8k & 452.8 & 10.7 & 1102.4 & 86.6 & 662.0 \\ 
    Wellness      & 16.0k & 396.5 & 10.5 & 2348.6 & 217.7 & 1025.8 \\ 
    Others        & 16.9k & 220.9 & 8.9 & 208.8 & 1266.2 & 3225.0 \\ 
    \bottomrule
    \end{tabular}}
    \caption{Statistics of the RedNote posts across content categories. The post count (\#) is presented in thousands (k). Comm. and Colls. refer to the average number of comments and collections, respectively.}
    \label{tab:domain_stats}
\end{table}

\subsection{Data Collection and Statistics}

Our data collection methodology is grounded in the \textit{2024 RedNote User Behavior Report} \footnote{http://xhslink.com/o/3ZtMUSbgoCl}, which identifies ten dominant content categories: \textit{Career, Wellness, Travel, Health, Food, Pets, Education, Sports, Fashion, and Relationships}. Our dataset focuses on the Chinese language. To ensure our dataset reflects real-world distribution, we employ a two-step retrieval strategy: We first extracted the official popular search keywords provided in the report for each category. For instance, the Relationships category includes specific seed tags such as ``Should couples go dutch?'' \begin{CJK*}{UTF8}{gbsn}
(情侣之间应该AA制吗)\end{CJK*} and ``First meeting with a blind date'' \begin{CJK*}{UTF8}{gbsn}
(相亲第一次见面)\end{CJK*}. Then, we expand these tag lists based on search recommendations, resulting in about 500 tags for different domains. We use these tags as search keywords to collect the notes. 

We collected 148,000 notes published between October 2020 and July 2025. To maintain the quality of the text content, we filter out notes with fewer than 100 Chinese characters, which were mainly image-based. Finally, our data source contains samples with comprehensive metadata including: 1) Content: note title, text content and tags; 2) Temporal information: publication timestamp; 3) Engagement metrics: likes, comments and collections; 4) Topic domain.

Table~\ref{tab:domain_stats} presents detailed statistics across domains, which shows that different domains exhibit distinct properties and engagement dynamics. Fig.~\ref{fig:dist}-(a) visualizes the distribution of the total engagement (defined as the sum of all three metrics), Fig.~\ref{fig:dist}-(b) presents the distribution of time span. It can be seen that our dataset covers different levels of interaction metrics, reflecting real-world user interaction behavior on social media.

\begin{table*}[h]
\centering
\small
\resizebox{\linewidth}{!}{%
\begin{tabular}{l|ccccccc}
\toprule
\textbf{Dataset} & \textbf{Data Source(s)} & \# & \textbf{Time Span} & \textbf{Timestamps} & \textbf{Language(s)} & \textbf{LLMs} & \textbf{Interact.} \\
\midrule
SM-D (\citeyear{sun2025we}) & Medium, Quora, Reddit & 2.4M & Jan 2022 - Oct 2024 & \ding{51} & English & 12 & \ding{51} \\
RAID (\citeyear{dugan2024raid}) & News, Books, Reddit, etc. & 6.2M & Not Applicable & \ding{53} & English & 11 & -- \\
MultiSocial (\citeyear{dugan2024raid}) & Twitter, WhatsApp, etc. & 472k & Mostly Pre-2022 & \ding{53} & Multi (22) & 7 & -- \\
MULTITuDE (\citeyear{macko2023multitude}) & News & 74k & Pre-2021 & \ding{53} & Multi (11) & 8 & -- \\
SAID (\citeyear{cui2023said}) & Zhihu & 72k & Mar 2023 - Oct 2023 & \ding{51} & Chinese & -- & -- \\
\midrule
RedNote-Vibe & RedNote & 148k & Oct 2020 - Jul 2025 & \ding{51} & Chinese & 18 & \ding{51} \\
\bottomrule
\end{tabular}
}
\caption{Comparison of RedNote-Vibe with prior social media AIGT datasets. Interact.: SM-D provides the number of followers, comments and likes.}
\label{tab:dataset_comparison}
\vspace{-0.3cm}
\end{table*}

\subsection{AI Samples Generation}

To create a comprehensive AIGT detection benchmark, we construct parallel AI-generated versions of RedNote posts. Unlike existing AIGT datasets that mainly direct prompt LLMs to generate AI samples, we employ a pipeline approach to simulate AIGT challenges in real-world scenarios. We design the generation method to be difficult for humans to distinguish. The detailed prompt is shown in Appendix~\ref{sec:ai_generation}.


We create a model pool comprising 17 representative LLMs from 6 providers. All models receive identical prompts with JSON-formatted output. Notably, we exclusively select seed notes published before November 2022 (pre-LLM period) to ensure their human-authored property. These seeds are randomly distributed across the LLM pool, with each model generating at least 1,000 samples.

Figure~\ref{fig:method} illustrates our data construction pipeline and the selected LLM pool. This approach yields a training and validation set with verified human/AI labels. Additionally, we compile an exploration set containing posts from the post-LLM period (2023-2025). While lacking ground-truth labels, this subset enables researchers to investigate real-world content evolution and analyze the emerging linguistic landscape shaped by widespread AI adoption.

In Table~\ref{tab:dataset_comparison}, we compare our dataset with existing publicly available datasets for AIGT detection in social media. While the existing datasets are valuable, RedNote-Vibe offers unique time spans and timestamps that support AIGT evolution and trend studies in Chinese context. Compared with most related SM-D \cite{sun2025we}, its sources are rooted in knowledge sharing and community discussions (e.g., Quora, Reddit), RedNote is a lifestyle-sharing platform integrated with commercial activities. This presents distinct challenges: AIGT on RedNote is often crafted into personal narratives (e.g., product reviews, travel experiences, etc.), making it significantly harder to detect compared to informational text. Furthermore, given the platform's commercial attributes, such content poses a more direct influence on consumer behavior. Therefore, RedNote-Vibe not only fills the gap of data source from the RedNote platform but also provides a novel testbed for studying AI-generated content in high-stakes, commercially driven social environments.

\subsection{Task Definition}

Leveraging our dataset's rich structure, we define three hierarchical classification tasks that reflect real-world AIGT detection scenarios with increasing granularity:

\begin{itemize}[itemsep=0pt, topsep=0pt, parsep=0pt]
\item \textbf{AIGT Classification (binary)}: Human vs. AI-generated text detection task, which requires models to distinguish human-written content from any AI-generated text. This task establishes the baseline capability for AIGT detection.

\item \textbf{AI Provider Identification (6-way)}: A task focusing on the AI-generated subset, where models identify the source among six major AI providers (OpenAI, Google, Anthropic, etc.). This task tests whether detection methods can capture family-level patterns, as models from the same provider often share similar training methodologies and corpus.

\item \textbf{Model Identification (17-way)}: A fine-grained AI model identification task, where we distinguish between 17 specific AI models within the AI-generated content, representing the most challenging scenario that requires detecting subtle model-specific characteristics.

\end{itemize}

\section{PsychoLinguistic AIGT Detection Framework }

\begin{table*}[t]
\centering
\small
\resizebox{\textwidth}{!}{
\begin{tabular}{p{0.25\textwidth} p{0.72\textwidth}}
\toprule
\textbf{Dimension} & \textbf{Features \& Theoretical Basis} \\
\midrule
\textbf{Emotional \& Social Grounding} \newline
\textit{Evaluates emotional depth, social connection, and sensory richness. rooted in personal experience and social awareness.} & 
$\bullet$ \textbf{Emotional Intensity} \citep{gross1998emerging}; \textbf{Personal Emotional Grounding} \citep{conway2000construction} \newline
$\bullet$ \textbf{Sensory Detail Richness} \citep{levine2002aging}; \textbf{Social Connectedness} \citep{giles1975speech} \newline
$\bullet$ \textbf{Empathetic Engagement} \citep{baron1997mindblindness}; \textbf{Interactive and Dialogic Stance} \citep{bakhtin2010dialogic} \newline
$\bullet$ \textbf{Unique Emoji Ratio} \citep{felbo2017using}; \textbf{Emoji Density} \citep{felbo2017using} \\
\midrule
\textbf{Cognitive Architecture} \newline
\textit{Assesses perspectival complexity, reasoning, and narrative structure.} & 
$\bullet$ \textbf{Perspectival Complexity} \citep{baker1992conceptual}; \textbf{Narrative Structure Flexibility} \citep{labov1997narrative} \newline
$\bullet$ \textbf{Dialectical Argumentation Strength} \citep{toulmin2003uses}; \textbf{Self-Correction} \citep{Hyland2005} \newline
$\bullet$ \textbf{Axiological Coherence} \citep{graham2009liberals}; \textbf{Temporal Orientation and Integration} \citep{reichenbach1947elements} \newline
$\bullet$ \textbf{Sentence/Word/Character Count} \citep{manning1999foundations} \\
\midrule
\textbf{Lexical Identity \& Stylistic Signature} \newline
\textit{Captures stylistic idiolect, rhythmic patterns, and natural imperfections.} & 
$\bullet$ \textbf{Lexical-Stylistic Personalization} \citep{argamon2003gender}; \textbf{Prosodic Rhythm Consistency} \citep{halliday2013halliday} \newline
$\bullet$ \textbf{Imperfection} \citep{clark2002using}; \textbf{Rhetorical Sophistication} \citep{grice1975logic} \newline
$\bullet$ \textbf{Punctuation Ratio}, \textbf{Number Ratio} \citep{manning1999foundations} \newline
$\bullet$ \textbf{Type-Token Ratio} \citep{templin1957certain}; \textbf{Word Frequency Entropy} \citep{shannon1948mathematical}; \textbf{Word Burstiness} \citep{gries2008dispersions} \\
\midrule
\textbf{Cohesion \& Textual Flow} \newline
\textit{Evaluates the organization, semantic progression, and repetition patterns.} & 
$\bullet$ \textbf{Lexical Cohesion} \citep{halliday2014cohesion}; \textbf{Inter-Sentential Sentence Similarity} \citep{foltz1998measurement} \newline
$\bullet$ \textbf{Immediate Repetition Density}, \textbf{Phrasal Repetition Frequency} \citep{stamatatos2009survey} \newline
$\bullet$ \textbf{Sentence Burstiness} \citep{gries2008dispersions} \\
\bottomrule
\end{tabular}
}
\caption{Feature list of the PLAD framework, categorized by psychological dimensions. Each feature is grounded in established psycholinguistic theories.}
\label{tab:full_feature_list}
\vspace{-0.3cm}
\end{table*}

Existing approaches to AIGT detection primarily rely on computational linguistic features or supervised training of PLMs, achieving significant performance. However, as AI continues to evolve, the linguistic features such as style and syntax that currently distinguish AIGT will inevitably be neutralized.
We hypothesize that detection methods grounded in established psycholinguistic dimensions (e.g., emotional expression patterns, narrative coherence, and lexical diversity) may be more robust to model updates, as these dimensions reflect fundamental differences in how text is generated. Consequently, we propose the PsychoLinguistic AIGT Detection Framework (PLAD), which leverages constructs from psycholinguistics as measurable features for interpretable and potentially more stable AIGT detection.


Our framework quantifies a total of 31 features categorized into four dimensions of human language expression: \textit{Emotional and Social Grounding}, \textit{Cognitive Architecture}, \textit{Lexical Identity and Stylistic Signature}, and \textit{Cohesion and Textual Flow}. These dimensions are rooted in psychological and cognitive theories, capturing aspects from emotional depth to narrative complexity. Table~\ref{tab:full_feature_list} presents the definition of each dimension and the corresponding features with their references.


Our feature set can be categorized into two extraction approaches. (1)  Directly computable statistical features that can be obtained using straightforward computational methods, such as emoji density, type-token ratio and other structural measures. (2) Semantically-based features that require evaluation criteria and text analysis tools for assessment. In contrast to traditional approaches to psychological text analysis such as LIWC \citep{pennebaker2015development}, which employs frequency-based word analysis, we adopt a more sophisticated approach using a proxy LLM for psychological text analysis.
Specifically, following related work \citep{rathje2024gpt, ghatora2024sentiment}, we design evaluation rubrics that convert theoretical constructs into measurable criteria for characteristics. These rubrics are then presented to a proxy LLM, which is instructed to evaluate the input text according to the specified dimensions and provide quantitative scores. 

Considering that using a proxy LLM for feature extraction may bring model biases into the feature set. However, we adopt this approach based on several reasons: (1) For complex psycholinguistic constructs, LLMs currently represent the most versatile and scalable assessment method; (2) Recent studies have demonstrated the robustness of LLM-based psychological assessment, and our validation of LLMs against human annotations aligns with previous results. The detailed implementation and evaluation are shown in Appendix~\ref{sec:impl}.

Based on the extracted feature vector $\mathbf{f}(x) \in \mathbb{R}^{31}$, we train a supervised classifier to predict the text's label. To ensure the framework's interpretability, we utilize tree-based models such as XGBoost and CatBoost, which provide clear feature importance rankings. The classification task is formally defined as finding the label $\hat{y} = \arg\max_{y \in \mathcal{Y}} P(y|\mathbf{f}(x))$, where $\mathcal{Y}$ is the set of possible labels. The model is trained by minimizing the cross-entropy loss, $\mathcal{L}_{\text{CE}} = -\sum_{i} y_i \log(p_i)$.

\section{Experiments}

\begin{table*}[!t]
\centering

\resizebox{0.95\textwidth}{!}{

\begin{tabular}{lccccccccc}
\toprule
\multirow{2}{*}{\textbf{Method}} & \multicolumn{3}{c}{\textbf{Model Identification (17-way)}} & \multicolumn{3}{c}{\textbf{Provider Identification (6-way)}} & \multicolumn{3}{c}{\textbf{AIGT Classification (binary)}} \\
\cmidrule(lr){2-4} \cmidrule(lr){5-7} \cmidrule(lr){8-10}
 & Precision & Recall & Acc. & Precision & Recall & Acc. & Precision & Recall & Acc. \\
\midrule
\multicolumn{10}{c}{\textbf{Statistics-based Methods}} \\
\midrule
StyloAI (\citeyear{opara2024styloai}) & 21.68 & 22.82 & 23.20 & 37.33 & 37.24 & 41.21 & 75.23 & 72.55  & 77.90 \\
\citet{ullah2024beyond}) & 23.83 & 25.29 & 26.01 & 41.05 & 41.04 & 46.41 & 75.91 & 73.83  & 78.59 \\
Binoculars (\citeyear{hans2024spotting}) & 21.13 & 22.67 & 24.05 & 40.48 & 39.14 & 43.72 & 72.07 & 74.17 & 77.89 \\
\midrule
\multicolumn{10}{c}{\textbf{Model-based Methods}} \\
\midrule
FT. BERT (\citeyear{devlin2019bert}) & 32.97 & 32.71 & 33.76 & 44.38 & 42.41 & 50.47 & 85.91 & 88.29 & 88.24 \\
FT. ALBERT (\citeyear{lan2019albert}) & 32.26 & 31.74 & 31.65 & 41.98 & 40.61 & 44.34 & 83.59 & 79.63 & 84.58 \\
FT. RoBERTa (\citeyear{liu2019roberta}) & \underline{35.20} & 31.78 & 32.21 & 38.22 & 40.20 & 47.64 & 87.85 & \underline{88.88} & \underline{89.52} \\
Sniffer (\citeyear{li2023origin}) & 30.63 & 30.77 & 30.04 & 44.39 & 42.88 & 45.65 & 82.55 & 81.30  & 84.45 \\
POGER (\citeyear{shi2024ten}) & 31.28 & 30.16 & 31.88 & 45.41 & 44.68 & 47.16 & 79.17 & 82.03  & 84.89 \\
LLM-Idiosyncrasies (\citeyear{sun2025idiosyncrasies}) & 32.91 & 33.16 & 32.31 & 49.81 & 45.71 & 49.96  & \underline{88.09} & \textbf{90.15}  & 89.07 \\

\midrule
\multicolumn{10}{c}{\textbf{PLAD Framework with Different Classifiers}} \\
\midrule
PLAD-XGBoost & 32.11 & \underline{33.51} & \underline{34.04} & \textbf{50.73} & \textbf{48.77} & \textbf{53.30} & 86.45 & 85.31 & 87.79 \\
PLAD-CatBoost & \textbf{35.87} & \textbf{36.45} & \textbf{36.94} & \underline{50.06} & \underline{47.34} & \underline{51.89} & \textbf{88.70} & 87.28 & \textbf{89.62} \\
\bottomrule
\end{tabular}}
\caption{Comparison of PLAD with existing methods across detection tasks derived from RedNote-Vibe. FT. denote fine-tuning. Precision, recall, and accuracy are computed as macro-averages. 
}
\label{model_comp}
\vspace{-0.3cm}

\end{table*}

\subsection{Benchmarking Detectors on RedNote-Vibe}

\textbf{Experiment Setup.} We compare AIGT detectors on our proposed tasks, with three categories of approaches. (1) \textbf{Statistics-based methods}: StyloAI~\citep{opara2024styloai}, Binoculars \citep{hans2024spotting} and the method of \citet{ullah2024beyond}, which serve as representative feature-driven baselines for AIGT detection. (2) \textbf{Model-based methods}: This category covers strong baselines, including (i) fine-tuning pre-trained text classification models such as BERT-base~\citep{devlin2019bert}, RoBERTa-base~\citep{liu2019roberta}, and ALBERT-base~\citep{lan2019albert}, which represent strong baselines~\citep{gritsai2024ai, li2024mage}, and (ii) state-of-the-art AIGT detection pipelines such as Sniffer~\citep{li2023origin}, POGER~\citep{shi2024ten}, and LLM-Idiosyncrasies~\citep{sun2025idiosyncrasies}. (3) \textbf{PLAD}: For comparison, we evaluate our proposed PLAD framework with different classifiers, i.e., CatBoost~\citep{prokhorenkova2018catboost} and XGBoost~\citep{chen2016xgboost}. We follow the original training protocols of all methods and implement them on our dataset. All methods share the same dataset and are split into training, validation, and testing sets in an 8:1:1 ratio. 

\textbf{Results.} Table~\ref{model_comp} summarizes the performance across three tasks. For \textit{AIGT classification}, competitive results are observed, with LLM-Idiosyncrasies achieving the highest recall while PLAD-CatBoost attains the best precision and accuracy. In the more granular \textit{provider} and \textit{model identification} tasks, PLAD consistently outperforms baselines. These results suggest that our psycholinguistic features effectively capture the subtle stylistic signatures of different LLMs, offering both superior detection capability and interpretability.


\subsection{Generalization to Unseen Models}

\begin{table}[t]
    \centering
    \small
    \begin{tabular}{clcc} 
    \toprule
     & \textbf{Method} & \textbf{Seen Acc.} & \textbf{Unseen Recall} \\ 
    \midrule
    \multirow{4}{*}{\rotatebox{90}{GPT-o3}} 
     & PLAD               & 46.46 & \textbf{56.34} \\
     & BERT-base          & 42.42 & 25.35 \\
     & ALBERT-base        & 42.73 & 28.31 \\
     & LLM-Idiosyncrasies & 44.34 & 22.54 \\
    \midrule
    \multirow{4}{*}{\rotatebox{90}{Gemini-2.5}} 
     & PLAD               & 43.72 & \textbf{58.46} \\
     & BERT-base          & 45.23 & 52.31 \\
     & ALBERT-base        & 42.61 & 50.00 \\
     & LLM-Idiosyncrasies & 41.71 & 53.85 \\
    \bottomrule
    \end{tabular}
    \caption{Recall on unseen target models (GPT-o3 and Gemini-2.5) in Provider Identification (6-way) task.}
    \label{tab:zero_shot_performance}
    \vspace{-0.2cm}
\end{table}
In this experiment, we evaluate the generalizability of detection methods on unseen AI models.
Given the observation that model providers often share similar training methodologies, data sources, and architectures across different versions. Therefore, a model family inherits a specific style imprint \citep{spiliopoulou2025play}. We design a zero-shot experiment by excluding the latest GPT-o3 and Gemini-2.5 from the training data while retaining other models from their respective providers (OpenAI and Google). This setup mimics the scenario that detection systems encounter newly released models that are not available during training. We set up experiments on the reduced dataset, then evaluate their accuracy on seen models and recall on unseen target models. 
As results are shown in Table~\ref{tab:zero_shot_performance}, our PLAD framework significantly outperforms other approaches in identifying unseen models, demonstrating that the psycholinguistic features extracted by PLAD can capture more robust and generalizable traces of model families. In the context of rapid iteration of LLMs, this detection capability of new models makes our framework more practical than model-based methods.

\subsection{Ablation and Feature Importance Study}

\begin{table}[b]
\centering
\small
    \begin{tabular}{lccc}
        \toprule
        \textbf{Configuration} & \textbf{17-way} & \textbf{6-way} & \textbf{binary} \\
        \midrule
        CatBoost (Full) & \textbf{36.16} & \textbf{48.66} & \textbf{87.98} \\
        w/o Emotional Dim.   & 27.68 & 43.22 & 86.32 \\
        w/o Cognitive Dim.   & 31.40 & 43.56 & 86.44 \\
        w/o Lexical Dim.  & 26.81 & 38.66 & 85.00 \\
        w/o Cohesion Dim. & 32.29 & 44.76 & 86.57 \\
        \bottomrule
    \end{tabular}
\caption{Ablation experiment result of each feature dimension. We report the macro-average F1-score (\%) on three classification tasks.}
\label{tab:ablation_study_f1}

\end{table}

\begin{figure*}[!t]
    \centering
    \includegraphics[width=0.99\linewidth]{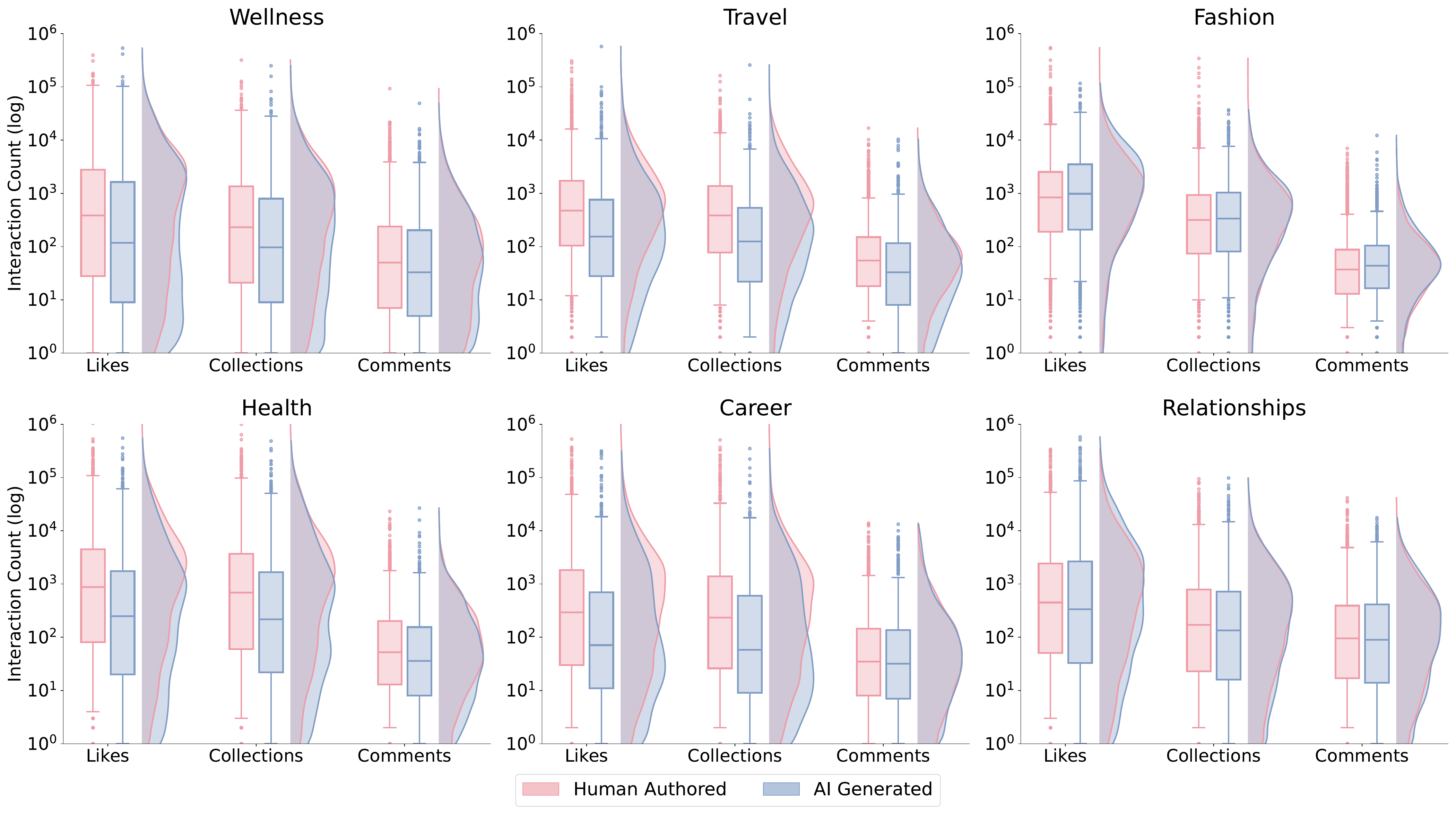}
    \caption{Engagement metrics comparison between Human-authored and AI-generated posts across 6 topical domains.}
    \label{fig:analysis-1}
    \vspace{-0.3cm}
\end{figure*}

To understand how PLAD achieves its detection performance, we conduct an analysis combining a dimension-level ablation study and evaluation of feature importance using CatBoost classifier.
Table~\ref{tab:ablation_study_f1} present the ablation study result. It reveals that the largest performance drop occurred when the Lexical Identity and Stylistic Signature features are removed, leading to a substantial decrease across all three tasks. We further identify the two most important features, \textit{Prosodic Rhythm Consistency} and \textit{Type-Token Ratio}, suggest that AI-generated text tends to exhibit smoother and more uniform rhythmic patterns, whereas human writing often contains irregularities caused by cognitive processing. The third most important feature, \textit{Imperfection}, also from this dimension, which detects the absence of human-like hesitations, self-corrections, and other disfluencies that LLMs are optimized to avoid. Detailed feature importance and the differences between human and AI text analysis are presented in Appendix~\ref{sec:feat_analysis}.


\section{Analysis}

In the previous experiments, we demonstrated that our proposed PLAD framework achieves state-of-the-art performance in detecting AIGT on the RedNote dataset, outperforming existing black-box baselines. We now apply PLAD to the real-world posts from the post-LLM era to investigate the temporal evolution of AIGT and its impact on social media engagement.

Considering that the exploration set (90,000 post-LLM posts) is inherently unlabeled, as only the authors themselves know whether AI was used, and human annotation is also infeasible. Therefore, we use pre-LLM posts to evaluate model reliability. We adopt binary AIGT classification and evaluate the model on pre-LLM validation set, which consists of posts written by humans and independent from the exploration set used in downstream analyses. To prioritize precision and minimize false positives, we set a conservative threshold of 0.9 based on validation performance, achieving a false positive rate (FPR) of 4.9\%. Note that this threshold is determined solely from the labeled validation set before any analysis is conducted. After that, we apply the model with the pre-determined threshold to the naturally unlabeled exploration set for subsequent analyses.
This process follows a validation-then-deployment pipeline and is consistent with previous AIGT analysis studies \citep{sun2025we, liang2025quantifying}.

\begin{figure}[t]
    \centering
    \includegraphics[width=0.45\textwidth]{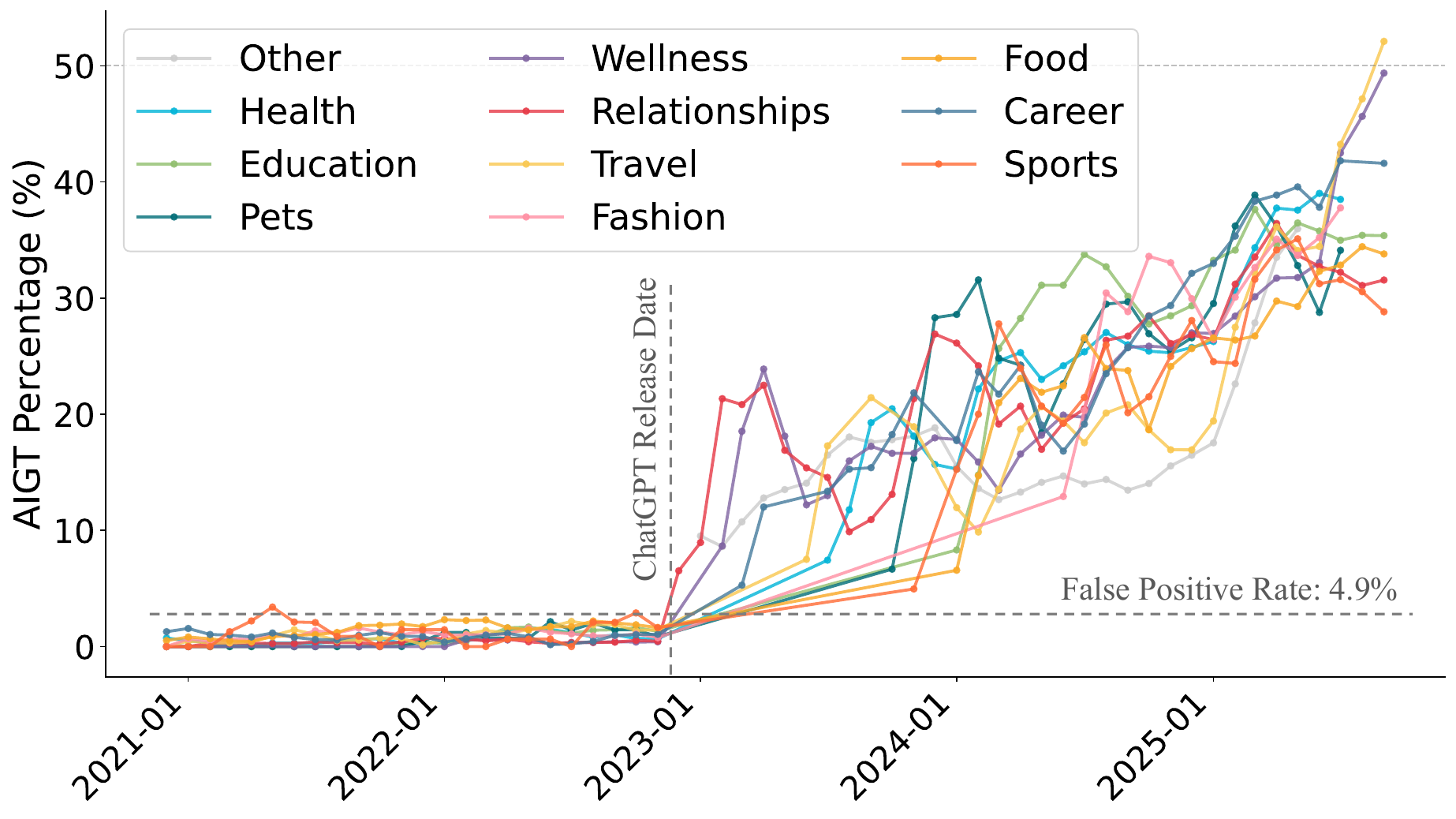}
    \caption{Temporal evolution of AIGT percentage across different domains (2021-2025).}
    \label{fig:trend}
    \vspace{-0.3cm}
\end{figure}

\subsection{Temporal Evolution of AI Content}

Figure~\ref{fig:trend} illustrates the monthly percentage of detected AI content across different domains. Since the emergence of LLMs, the proportion of AI-generated content has exhibited a steady growth across all domains. This trend indicates that \textbf{AI tools are increasingly utilized for lifestyle sharing}. This finding aligns with recent studies on other datasets \cite{sun2025we, liang2025quantifying}, reinforcing the observation that the adoption of generative AI is a widespread phenomenon.

\subsection{Metrics of AI vs. Human Posts}

To investigate the impact of content source on interaction, we compare engagement metrics across six representative domains in Figure~\ref{fig:analysis-1}.
It reveals that \textbf{Human content exhibits superior emotional resonance.}
Human-authored posts consistently outperform AI-generated ones in median engagement. This gap is particularly visible in domains requiring personal experience and emotional resonance, such as \textit{Travel}, \textit{Career}, and \textit{Relationships}.

Compared to human content, the interaction distribution for AI posts is more concentrated, with notably fewer high-engagement outliers. This indicates that while AI can generate grammatically correct text, it lacks the novelty and high-arousal qualities necessary to create breakout content.
Looking into the variations between interaction types reveals the cognitive investment from users. For instance, a like is a passive and lightweight social signal, whereas a comment requires active and conscious engagement. Interestingly, the human-AI gap in comments is smaller than in likes. This suggests that \textbf{while AI content receives lower overall engagement, the performance gap narrows for interactions requiring higher cognitive investment}.

\subsection{Author-Level AI Usage and Engagement}

\begin{figure}[b]

    \centering
    \includegraphics[width=0.45\textwidth]{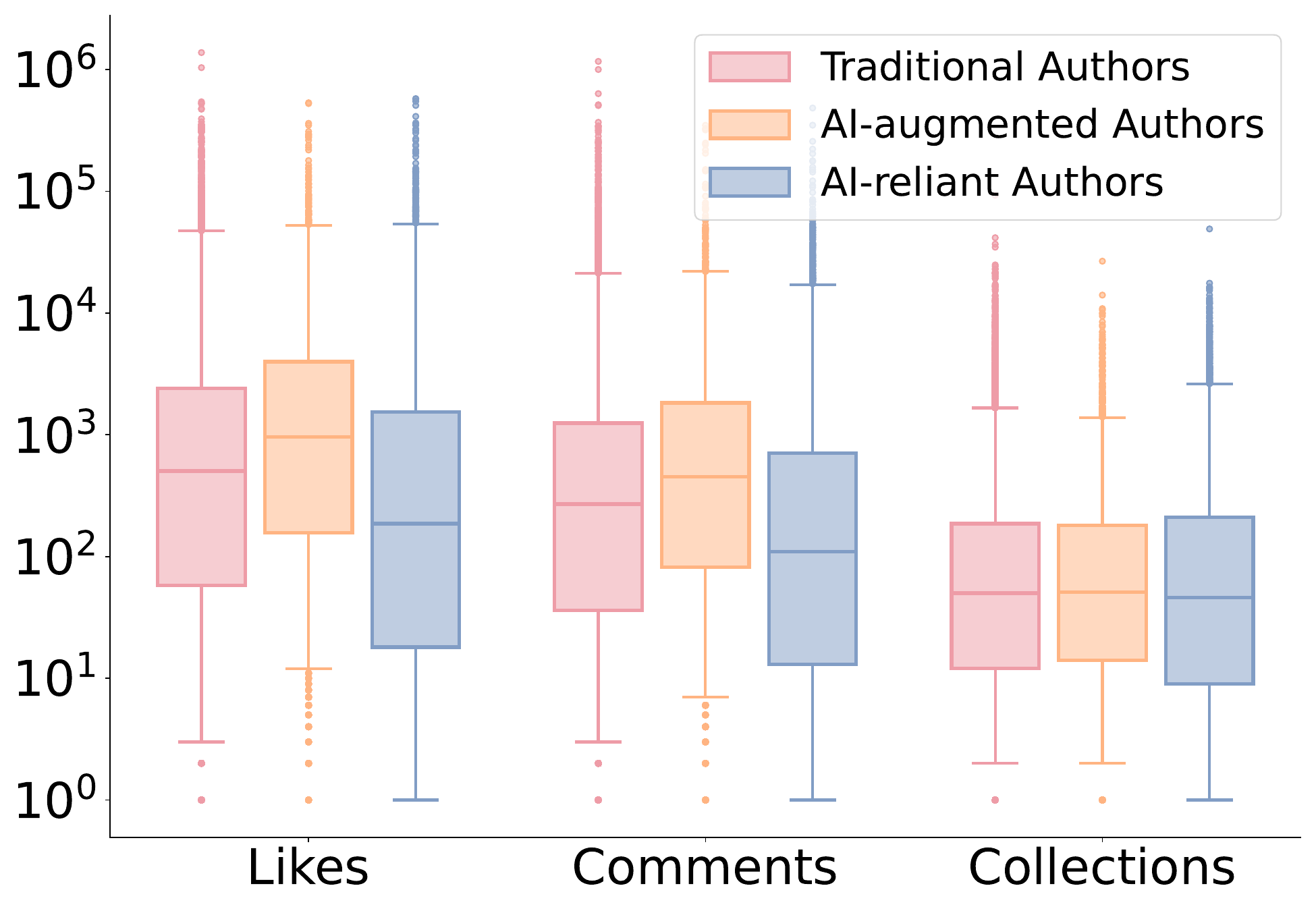}
    \caption{Author-level analysis of AI usage patterns and engagement outcomes.}
    \label{fig:analysis-3}
    
\end{figure}

In addition to content-level analysis, we conduct an author-centric study to understand how different AI usage strategies correlate with user interaction behavior. We select authors with at least four posts in our dataset to estimate their AI usage, resulting in 829 authors. We compute each author's AI ratio by their posts classified as AIGT by PLAD. Based on this, we categorize authors into three groups: (1) \textit{Traditional Authors} (AI ratio = 0\%), who compose entirely without AI assistance (570 authors, 68.7\%); (2) \textit{AI-reliant Authors} (AI ratio = 100\%), who fully rely on AI for all their posts (230 authors, 27.7\%); and (3) \textit{AI-augmented Authors} (0\% $<$ AI ratio $<$ 100\%), who strategically combine human and AI-generated content (29 authors, 3.5\%). For each author group, we present their engagement metrics in Figure~\ref{fig:analysis-3}.

The results of Traditional Author and AI-reliant Author groups are consistent with the findings in the previous section. However, the AI-augmented Authors group achieves relatively higher engagement. We hypothesize that these authors leverage AI for efficiency while applying human judgment to refine the content quality. This might suggest that \textbf{a small group of users who strategically utilize AI tools can achieve higher engagement outcomes}, outperforming both those who avoid AI entirely and those who rely on it exclusively.

\section{Discussion and Conclusion}

\noindent\textbf{Rethinking AIGT Detection in the Wild.}
Existing detection methods often report high accuracy (e.g., $>98\%$). However, we argue that this performance is inflated by the stylistic gap between naive LLM outputs and human writing. 
In this work, we optimized our generation pipeline to produce content more indistinguishable from human posts. It makes RedNote-Vibe more challenging, reflecting real-world adversarial conditions.

\noindent\textbf{Longitudinal Interactions Between Human and AI Content.}
As AIGT becomes increasingly prevalent on social media, users are exposed to both human- and AI-generated content with growing frequency. LLMs are trained on human-produced corpora, while their outputs may subsequently influence human writing patterns. Our dataset provides a resource for investigating these temporal dynamics, which we leave for future work.

\noindent\textbf{Conclusion.}
In this paper, we introduce a novel dataset that fills the gap in Chinese lifestyle social media resources. We also propose an AIGT detection method as the baseline for our dataset and tasks. Our exploratory analysis reveals some AIGT and user behavior patterns of social media. We emphasize that our core contribution lies in the dataset itself, we believe it provides a valuable data source for future AI/human behavior research.



\section*{Ethical Statement}
All data utilized in this study originates from publicly accessible posts on the RedNote platform. We do not collect any private information. For the author-centric study, we hashed the usernames for grouping, and we only performed statistical analysis, not on individuals. 

Regarding the AI-generated variants, they are produced using commercial LLM APIs with the sole objective of constructing a robust detection benchmark. We explicitly prohibit the misuse of this dataset for generating misleading content or spam. 

\section*{Limitations}
While our work provides comprehensive insights into AIGT on social media, we acknowledge several limitations that future work should address:

\noindent\textbf{Language and Platform Specificity.}
Our study focuses exclusively on RedNote (Xiaohongshu), a platform with dominant of Chinese-language users. Consequently, the linguistic features and engagement patterns observed here may not directly generalize to other platforms. Furthermore, although our PLAD framework is designed based on general psycholinguistic principles and is theoretically language-agnostic, we have not empirically verified its efficacy across other languages in this study.

\noindent\textbf{Limited Coverage of Generative Models.}
The AIGT variants in our dataset are generated using a subset of LLMs available up to mid-2025. Given the rapid iteration of model capabilities (e.g., the emergence of reasoning-heavy models), the generated vibe may continue to evolve. 

\noindent\textbf{Reliance on Proxy LLMs.}
Our feature extraction pipeline relies on a proxy LLM to quantify psychological dimensions. Biases present in the proxy LLM could propagate into the feature set. However, these biases are unlikely to significantly impact the analysis results. In the Appendix~\ref{sec:reliability_of_proxy_llms}, we provide a robustness evaluation, demonstrating that our method outperforms existing approaches in terms of correlation with human judgments.

\noindent\textbf{Cost of Inference.}
Compared to lightweight end-to-end classifiers (e.g., RoBERTa-based detectors), PLAD involves a multi-step process of feature extraction via LLMs, which incurs higher computational and financial costs. This may limit its real-time application in high-throughput industrial scenarios without further optimization.


\bibliography{custom}

@article{sun2025idiosyncrasies,
  title={Idiosyncrasies in large language models},
  author={Sun, Mingjie and Yin, Yida and Xu, Zhiqiu and Kolter, J Zico and Liu, Zhuang},
  journal={arXiv preprint arXiv:2502.12150},
  year={2025}
}

@article{hu2023radar,
  title={Radar: Robust ai-text detection via adversarial learning},
  author={Hu, Xiaomeng and Chen, Pin-Yu and Ho, Tsung-Yi},
  journal={Advances in neural information processing systems},
  volume={36},
  pages={15077--15095},
  year={2023}
}

@article{spiliopoulou2025play,
  title={Play Favorites: A Statistical Method to Measure Self-Bias in LLM-as-a-Judge},
  author={Spiliopoulou, Evangelia and Fogliato, Riccardo and Burnsky, Hanna and Soliman, Tamer and Ma, Jie and Horwood, Graham and Ballesteros, Miguel},
  journal={arXiv preprint arXiv:2508.06709},
  year={2025}
}

@article{gritsai2024ai,
  title={Are ai detectors good enough? a survey on quality of datasets with machine-generated texts},
  author={Gritsai, German and Voznyuk, Anastasia and Grabovoy, Andrey and Chekhovich, Yury},
  journal={arXiv preprint arXiv:2410.14677},
  year={2024}
}

@inproceedings{li2024mage,
  title={MAGE: Machine-generated Text Detection in the Wild},
  author={Li, Yafu and Li, Qintong and Cui, Leyang and Bi, Wei and Wang, Zhilin and Wang, Longyue and Yang, Linyi and Shi, Shuming and Zhang, Yue},
  booktitle={Proceedings of the 62nd Annual Meeting of the Association for Computational Linguistics (Volume 1: Long Papers)},
  pages={36--53},
  year={2024}
}

@article{pennebaker2015development,
  title={The development and psychometric properties of LIWC2015},
  author={Pennebaker, James W and Boyd, Ryan L and Jordan, Kayla and Blackburn, Kate},
  year={2015}
}

@article{rathje2024gpt,
  title={GPT is an effective tool for multilingual psychological text analysis},
  author={Rathje, Steve and Mirea, Dan-Mircea and Sucholutsky, Ilia and Marjieh, Raja and Robertson, Claire E and Van Bavel, Jay J},
  journal={Proceedings of the National Academy of Sciences},
  volume={121},
  number={34},
  pages={e2308950121},
  year={2024},
  publisher={National Academy of Sciences}
}

@article{ghatora2024sentiment,
  title={Sentiment analysis of product reviews using machine learning and pre-trained llm},
  author={Ghatora, Pawanjit Singh and Hosseini, Seyed Ebrahim and Pervez, Shahbaz and Iqbal, Muhammad Javed and Shaukat, Nabil},
  journal={Big Data and Cognitive Computing},
  volume={8},
  number={12},
  pages={199},
  year={2024},
  publisher={MDPI}
}

@inproceedings{gui2025aider,
  title={AIDER: A robust and topic-independent framework for detecting AI-generated text},
  author={Gui, Jiayi and Cui, Baitong and Guo, Xiaolian and Yu, Ke and Wu, Xiaofei},
  booktitle={Proceedings of the 31st international conference on computational linguistics},
  pages={9299--9310},
  year={2025}
}

@article{guo2025deepseek,
  title={Deepseek-r1: Incentivizing reasoning capability in llms via reinforcement learning},
  author={Guo, Daya and Yang, Dejian and Zhang, Haowei and Song, Junxiao and Zhang, Ruoyu and Xu, Runxin and Zhu, Qihao and Ma, Shirong and Wang, Peiyi and Bi, Xiao and others},
  journal={arXiv preprint arXiv:2501.12948},
  year={2025}
}

@article{achiam2023gpt,
	title={Gpt-4 technical report},
	author={Achiam, Josh and Adler, Steven and Agarwal, Sandhini and Ahmad, Lama and Akkaya, Ilge and Aleman, Florencia Leoni and Almeida, Diogo and Altenschmidt, Janko and Altman, Sam and Anadkat, Shyamal and others},
	journal={arXiv preprint arXiv:2303.08774},
	year={2023}
}

@misc{google2025gemini2,
	author       = {Shrestha Basu Mallick and Logan Kilpatrick},
	title        = {Gemini 2.0:Flash, Flash-Lite and Pro},
	year         = {2025},
	month        = {February},
	url          = {https://developers.googleblog.com/zh-hans/gemini-2-family-expands/},
	note         = {Accessed: 2025-05-01}
}

@article{fagni2021tweepfake,
  title={TweepFake: About detecting deepfake tweets},
  author={Fagni, Tiziano and Falchi, Fabrizio and Gambini, Margherita and Martella, Antonio and Tesconi, Maurizio},
  journal={Plos one},
  volume={16},
  number={5},
  pages={e0251415},
  year={2021},
  publisher={Public Library of Science San Francisco, CA USA}
}

@article{schafer2024electionrumors2022,
  title={ElectionRumors2022: A dataset of election rumors on Twitter during the 2022 US midterms},
  author={Schafer, Joseph S and Duskin, Kayla and Prochaska, Stephen and Wack, Morgan and Beers, Anna and Bozarth, Lia and Agajanian, Taylor and Caulfield, Mike and Spiro, Emma S and Starbird, Kate},
  journal={arXiv preprint arXiv:2407.16051},
  year={2024}
}

@inproceedings{peng2015named,
title={Named entity recognition for chinese social media with jointly trained embeddings},
author={Peng, Nanyun and Dredze, Mark},
booktitle={Proceedings of the 2015 conference on empirical methods in natural language processing},
pages={548--554},
year={2015}
}

@article{yang2021checked,
  title={CHECKED: Chinese COVID-19 fake news dataset},
  author={Yang, Chen and Zhou, Xinyi and Zafarani, Reza},
  journal={Social Network Analysis and Mining},
  volume={11},
  number={1},
  pages={58},
  year={2021},
  publisher={Springer}
}

@inproceedings{kirchenbauer2023watermark,
  title={A watermark for large language models},
  author={Kirchenbauer, John and Geiping, Jonas and Wen, Yuxin and Katz, Jonathan and Miers, Ian and Goldstein, Tom},
  booktitle={International Conference on Machine Learning},
  pages={17061--17084},
  year={2023},
  organization={PMLR}
}

@article{liu2024does,
  title={Does detectgpt fully utilize perturbation? bridging selective perturbation to fine-tuned contrastive learning detector would be better},
  author={Liu, Shengchao and Liu, Xiaoming and Wang, Yichen and Cheng, Zehua and Li, Chengzhengxu and Zhang, Zhaohan and Lan, Yu and Shen, Chao},
  journal={arXiv preprint arXiv:2402.00263},
  year={2024}
}

@inproceedings{devlin2019bert,
  title={Bert: Pre-training of deep bidirectional transformers for language understanding},
  author={Devlin, Jacob and Chang, Ming-Wei and Lee, Kenton and Toutanova, Kristina},
  booktitle={Proceedings of the 2019 conference of the North American chapter of the association for computational linguistics: human language technologies, volume 1 (long and short papers)},
  pages={4171--4186},
  year={2019}
}

@article{liu2019roberta,
  title={Roberta: A robustly optimized bert pretraining approach},
  author={Liu, Yinhan and Ott, Myle and Goyal, Naman and Du, Jingfei and Joshi, Mandar and Chen, Danqi and Levy, Omer and Lewis, Mike and Zettlemoyer, Luke and Stoyanov, Veselin},
  journal={arXiv preprint arXiv:1907.11692},
  year={2019}
}

@article{huang2024ai,
  title={Are AI-Generated Text Detectors Robust to Adversarial Perturbations?},
  author={Huang, Guanhua and Zhang, Yuchen and Li, Zhe and You, Yongjian and Wang, Mingze and Yang, Zhouwang},
  booktitle={Proceedings of the 62nd Annual Meeting of the Association for Computational Linguistics (Volume 1: Long Papers)},
  pages={6005--6024},
  year={2024}
}

@inproceedings{mitchell2023detectgpt,
  title={Detectgpt: Zero-shot machine-generated text detection using probability curvature},
  author={Mitchell, Eric and Lee, Yoonho and Khazatsky, Alexander and Manning, Christopher D and Finn, Chelsea},
  booktitle={International Conference on Machine Learning},
  pages={24950--24962},
  year={2023},
  organization={PMLR}
}

@inproceedings{zhang2024machine,
  title={Machine-Generated Text Localization},
  author={Zhang, Zhongping and Qin, Wenda and Plummer, Bryan},
  booktitle={Findings of the Association for Computational Linguistics ACL 2024},
  pages={8357--8371},
  year={2024}
}

@inproceedings{opara2024styloai,
  title={StyloAI: Distinguishing AI-Generated Content with Stylometric Analysis},
  author={Opara, Chidimma},
  booktitle={25th International Conference on Artificial on Artificial Intelligence in Education},
  year={2024}
}

@article{cascio2024high,
  title={How high-arousal language shapes micro-versus macro-influencers’ impact},
  author={Cascio Rizzo, Giovanni Luca and Villarroel Ordenes, Francisco and Pozharliev, Rumen and De Angelis, Matteo and Costabile, Michele},
  journal={Journal of Marketing},
  volume={88},
  number={4},
  pages={107--128},
  year={2024},
  publisher={SAGE Publications Sage CA: Los Angeles, CA}
}

@article{chung2023really, 
  title={I really know you: How influencers can increase audience engagement by referencing their close social ties},
  author={Chung, Jaeyeon and Ding, Yu and Kalra, Ajay},
  journal={Journal of Consumer Research},
  volume={50},
  number={4},
  pages={683--703},
  year={2023},
  publisher={Oxford University Press}
}

@inproceedings{ullah2024beyond,
  title={Beyond Words: Stylometric Analysis for Detecting AI Manipulation on Social Media},
  author={Ullah, Ubaid and Laudanna, Sonia and Vinod, P and Di Sorbo, Andrea and Visaggio, Corrado Aaron and Canfora, Gerardo},
  booktitle={European Symposium on Research in Computer Security},
  pages={208--228},
  year={2024},
  organization={Springer}
}

@inproceedings{hans2024spotting,
  title={Spotting LLMs with binoculars: zero-shot detection of machine-generated text},
  author={Hans, Abhimanyu and Schwarzschild, Avi and Cherepanova, Valeriia and Kazemi, Hamid and Saha, Aniruddha and Goldblum, Micah and Geiping, Jonas and Goldstein, Tom},
  booktitle={Proceedings of the 41st International Conference on Machine Learning},
  pages={17519--17537},
  year={2024}
}

@article{li2023origin,
  title={Origin tracing and detecting of llms},
  author={Li, Linyang and Wang, Pengyu and Ren, Ke and Sun, Tianxiang and Qiu, Xipeng},
  journal={arXiv preprint arXiv:2304.14072},
  year={2023}
}

@article{shi2024ten,
  title={Ten words only still help: Improving black-box ai-generated text detection via proxy-guided efficient re-sampling},
  author={Shi, Yuhui and Sheng, Qiang and Cao, Juan and Mi, Hao and Hu, Beizhe and Wang, Danding},
  journal={arXiv preprint arXiv:2402.09199},
  year={2024}
}

@inproceedings{qiu2024paircfr,
  title={PairCFR: Enhancing Model Training on Paired Counterfactually Augmented Data through Contrastive Learning},
  author={Qiu, Xiaoqi and Wang, Yongjie and Guo, Xu and Zeng, Zhiwei and Yue, Yu and Feng, Yuhong and Miao, Chunyan},
  booktitle={Proceedings of the 62nd Annual Meeting of the Association for Computational Linguistics (Volume 1: Long Papers)},
  pages={11955--11971},
  year={2024}
}

@article{sadasivan2023can,
  title={Can AI-generated text be reliably detected?},
  author={Sadasivan, Vinu Sankar and Kumar, Aounon and Balasubramanian, Sriram and Wang, Wenxiao and Feizi, Soheil},
  journal={arXiv preprint arXiv:2303.11156},
  year={2023}
}

@article{lan2019albert,
  title={Albert: A lite bert for self-supervised learning of language representations},
  author={Lan, Zhenzhong and Chen, Mingda and Goodman, Sebastian and Gimpel, Kevin and Sharma, Piyush and Soricut, Radu},
  journal={arXiv preprint arXiv:1909.11942},
  year={2019}
}

@article{prokhorenkova2018catboost,
  title={CatBoost: unbiased boosting with categorical features},
  author={Prokhorenkova, Liudmila and Gusev, Gleb and Vorobev, Aleksandr and Dorogush, Anna Veronika and Gulin, Andrey},
  journal={Advances in neural information processing systems},
  volume={31},
  year={2018}
}

@inproceedings{chen2016xgboost,
  title={Xgboost: A scalable tree boosting system},
  author={Chen, Tianqi and Guestrin, Carlos},
  booktitle={Proceedings of the 22nd acm sigkdd international conference on knowledge discovery and data mining},
  pages={785--794},
  year={2016}
}

@article{gross1998emerging,
  title={The emerging field of emotion regulation: An integrative review},
  author={Gross, James J},
  journal={Review of general psychology},
  volume={2},
  number={3},
  pages={271--299},
  year={1998},
  publisher={SAGE Publications Sage CA: Los Angeles, CA}
}

@article{conway2000construction,
  title={The construction of autobiographical memories in the self-memory system.},
  author={Conway, Martin A and Pleydell-Pearce, Christopher W},
  journal={Psychological review},
  volume={107},
  number={2},
  pages={261},
  year={2000},
  publisher={American Psychological Association}
}

@article{levine2002aging,
  title={Aging and autobiographical memory: dissociating episodic from semantic retrieval.},
  author={Levine, Brian and Svoboda, Eva and Hay, Janine F and Winocur, Gordon and Moscovitch, Morris},
  journal={Psychology and aging},
  volume={17},
  number={4},
  pages={677},
  year={2002},
  publisher={American Psychological Association}
}

@book{giles1975speech,
  title={Speech style and social evaluation.},
  author={Giles, Howard and Powesland, Peter F},
  year={1975},
  publisher={Academic Press}
}

@book{baron1997mindblindness,
  title={Mindblindness: An essay on autism and theory of mind},
  author={Baron-Cohen, Simon},
  year={1997},
  publisher={MIT press}
}

@book{bakhtin2010dialogic,
  title={The dialogic imagination: Four essays},
  author={Bakhtin, Mikhail Mikhauilovich},
  volume={1},
  year={2010},
  publisher={University of texas Press}
}

@article{felbo2017using,
  title={Using millions of emoji occurrences to learn any-domain representations for detecting sentiment, emotion and sarcasm},
  author={Felbo, Bjarke and Mislove, Alan and S{\o}gaard, Anders and Rahwan, Iyad and Lehmann, Sune},
  journal={arXiv preprint arXiv:1708.00524},
  year={2017}
}

@article{baker1992conceptual,
  title={The conceptual/integrative complexity scoring manual},
  author={Baker-Brown, Gloria and Ballard, Elizabeth J and Bluck, Susan and De Vries, Brian and Suedfeld, Peter and Tetlock, Philip E},
  journal={Motivation and personality: Handbook of thematic content analysis},
  pages={401--418},
  year={1992}
}

@article{labov1997narrative,
  title={Narrative analysis: Oral versions of personal experience.},
  author={Labov, William and Waletzky, Joshua},
  year={1997},
  publisher={John Benjamins}
}

@inproceedings{macko2023multitude,
  title={MULTITuDE: Large-scale multilingual machine-generated text detection benchmark},
  author={Macko, Dominik and Moro, Robert and Uchendu, Adaku and Lucas, Jason and Yamashita, Michiharu and Pikuliak, Mat{\'u}{\v{s}} and Srba, Ivan and Le, Thai and Lee, Dongwon and Simko, Jakub and others},
  booktitle={Proceedings of the 2023 Conference on Empirical Methods in Natural Language Processing},
  pages={9960--9987},
  year={2023}
}

@book{toulmin2003uses,
  title={The uses of argument},
  author={Toulmin, Stephen E},
  year={2003},
  publisher={Cambridge university press}
}

@article{graham2009liberals,
  title={Liberals and conservatives rely on different sets of moral foundations.},
  author={Graham, Jesse and Haidt, Jonathan and Nosek, Brian A},
  journal={Journal of personality and social psychology},
  volume={96},
  number={5},
  pages={1029},
  year={2009},
  publisher={American Psychological Association}
}

@article{reichenbach1947elements,
  title={Elements of symbolic logic},
  author={Reichenbach, Hans},
  year={1947}
}

@book{Hyland2005,
  author    = {Hyland, Ken},
  title     = {Metadiscourse: Exploring Interaction in Writing},
  publisher = {Continuum},
  year      = {2005},
  address   = {London}
}

@book{manning1999foundations,
  title={Foundations of statistical natural language processing},
  author={Manning, Christopher and Schutze, Hinrich},
  year={1999},
  publisher={MIT press}
}

@article{argamon2003gender,
  title={Gender, genre, and writing style in formal written texts},
  author={Argamon, Shlomo and Koppel, Moshe and Fine, Jonathan and Shimoni, Anat Rachel},
  journal={Text \& talk},
  volume={23},
  number={3},
  pages={321--346},
  year={2003},
  publisher={De Gruyter Berlin, Germany}
}

@book{halliday2013halliday,
  title={Halliday's introduction to functional grammar},
  author={Halliday, Michael Alexander Kirkwood and Matthiessen, Christian MIM},
  year={2013},
  publisher={Routledge}
}

@article{clark2002using,
  title={Using uh and um in spontaneous speaking},
  author={Clark, Herbert H and Tree, Jean E Fox},
  journal={Cognition},
  volume={84},
  number={1},
  pages={73--111},
  year={2002},
  publisher={Elsevier}
}

@incollection{grice1975logic,
  title={Logic and conversation},
  author={Grice, Herbert P},
  booktitle={Speech acts},
  pages={41--58},
  year={1975},
  publisher={Brill}
}

@article{templin1957certain,
  title={Certain language skills in children; their development and interrelationships.},
  author={Templin, Mildred C},
  year={1957},
  publisher={University of Minnesota Press}
}

@article{shannon1948mathematical,
  title={A mathematical theory of communication},
  author={Shannon, Claude E},
  journal={The Bell system technical journal},
  volume={27},
  number={3},
  pages={379--423},
  year={1948},
  publisher={Nokia Bell Labs}
}

@article{gries2008dispersions,
  title={Dispersions and adjusted frequencies in corpora},
  author={Gries, Stefan Th},
  journal={International journal of corpus linguistics},
  volume={13},
  number={4},
  pages={403--437},
  year={2008},
  publisher={John Benjamins}
}

@book{halliday2014cohesion,
  title={Cohesion in english},
  author={Halliday, Michael Alexander Kirkwood and Hasan, Ruqaiya},
  year={2014},
  publisher={Routledge}
}

@article{foltz1998measurement,
  title={The measurement of textual coherence with latent semantic analysis},
  author={Foltz, Peter W and Kintsch, Walter and Landauer, Thomas K},
  journal={Discourse processes},
  volume={25},
  number={2-3},
  pages={285--307},
  year={1998},
  publisher={Taylor \& Francis}
}

@article{stamatatos2009survey,
  title={A survey of modern authorship attribution methods},
  author={Stamatatos, Efstathios},
  journal={Journal of the American Society for information Science and Technology},
  volume={60},
  number={3},
  pages={538--556},
  year={2009},
  publisher={Wiley Online Library}
}

@article{lundberg2017unified,
  title={A unified approach to interpreting model predictions},
  author={Lundberg, Scott M and Lee, Su-In},
  journal={Advances in neural information processing systems},
  volume={30},
  year={2017}
}

@inproceedings{sun2025we,
  title={Are we in the AI-generated text world already? Quantifying and monitoring AIGT on social media},
  author={Sun, Zhen and Zhang, Zongmin and Shen, Xinyue and Zhang, Ziyi and Liu, Yule and Backes, Michael and Zhang, Yang and He, Xinlei},
  booktitle={Proceedings of the 63rd Annual Meeting of the Association for Computational Linguistics (Volume 1: Long Papers)},
  pages={22975--23005},
  year={2025}
}

@article{cui2023said,
  title={Who said that? benchmarking social media ai detection},
  author={Cui, Wanyun and Zhang, Linqiu and Wang, Qianle and Cai, Shuyang},
  journal={arXiv preprint arXiv:2310.08240},
  year={2023}
}

@inproceedings{dugan2024raid,
  title={Raid: A shared benchmark for robust evaluation of machine-generated text detectors},
  author={Dugan, Liam and Hwang, Alyssa and Trhl{\'\i}k, Filip and Zhu, Andrew and Ludan, Josh Magnus and Xu, Hainiu and Ippolito, Daphne and Callison-Burch, Chris},
  booktitle={Proceedings of the 62nd Annual Meeting of the Association for Computational Linguistics (Volume 1: Long Papers)},
  pages={12463--12492},
  year={2024}
}

@article{liang2025quantifying,
  title={Quantifying large language model usage in scientific papers},
  author={Liang, Weixin and Zhang, Yaohui and Wu, Zhengxuan and Lepp, Haley and Ji, Wenlong and Zhao, Xuandong and Cao, Hancheng and Liu, Sheng and He, Siyu and Huang, Zhi and others},
  journal={Nature Human Behaviour},
  pages={1--11},
  year={2025},
  publisher={Nature Publishing Group UK London}
}
\clearpage

\appendix

\newpage
\newpage

\section*{\textbf{Appendix}}

In the Appendix, we provide related work (Section~\ref{rl}) and further analysis, including the differences between human-written text and AI-generated text from the perspective of the PLAD framework (Section~\ref{sec:feat_analysis-1}) and how these features affect the engagement metrics of posts (Section~\ref{sec:feat_analysis-2}). In addition, we analyze the correlation between features, reflecting the orthogonality of our framework (Section~\ref{sec:feat_analysis-3}). Finally, we introduce details of our AI samples generation (Section~\ref{sec:ai_generation}), feature extraction method (Section~\ref{sec:feat_extraction}) and the validation of our feature quality (Section~\ref{sec:reliability_of_proxy_llms}).

\section{Related Work}
\label{rl}

Social media datasets have been developed for AIGC detection, including TweepFake \citep{fagni2021tweepfake} for early Twitter content, SAID \citep{cui2023said} for modern LLM detection, and ElectionRumors2022 \citep{schafer2024electionrumors2022} for election-related content analysis. Chinese social media datasets focus primarily on Sina Weibo for tasks like NER \citep{peng2015named} and fake news detection \citep{yang2021checked}, but lack coverage of RedNote platform and temporal dynamics.

AIGC detection methods span three categories: watermarking techniques that embed imperceptible marks during generation \citep{kirchenbauer2023watermark,liu2024does}, classifier-based approaches using fine-tuned transformers like BERT and RoBERTa \citep{hu2023radar,huang2024ai}, and statistical methods that establish discrimination thresholds \citep{mitchell2023detectgpt,zhang2024machine}. Multi-class detection frameworks like Sniffer \citep{li2023origin} and LLM-Idiosyncrasies \citep{sun2025idiosyncrasies} leverage LLM embeddings for classification. However, existing statistical methods struggle with real-world accuracy \citep{qiu2024paircfr,sadasivan2023can}, highlighting the need for more robust and interpretable approaches.

\section{Further Analysis}
\label{sec:feat_analysis}

\subsection{Feature Statistics for Human and AI Post}
\label{sec:feat_analysis-1}

\begin{figure}[h]
\centering
\includegraphics[width=0.5\textwidth]{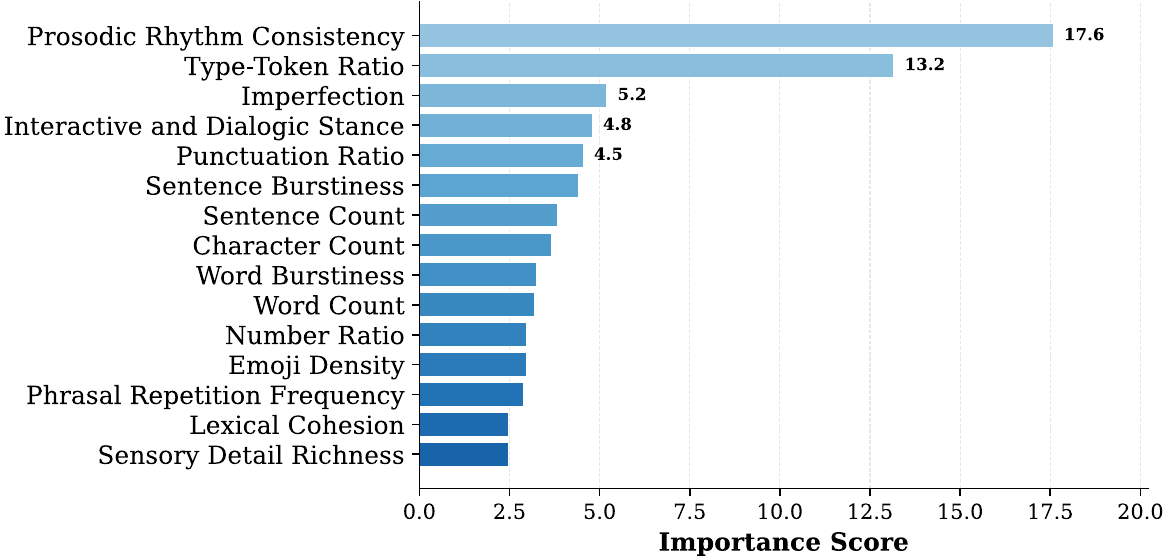}
\caption{The top-15 important feature scores.}
\label{fig:feat_imp}

\end{figure}

\begin{figure*}[!b]
    \centering
    \includegraphics[width=0.8\linewidth]{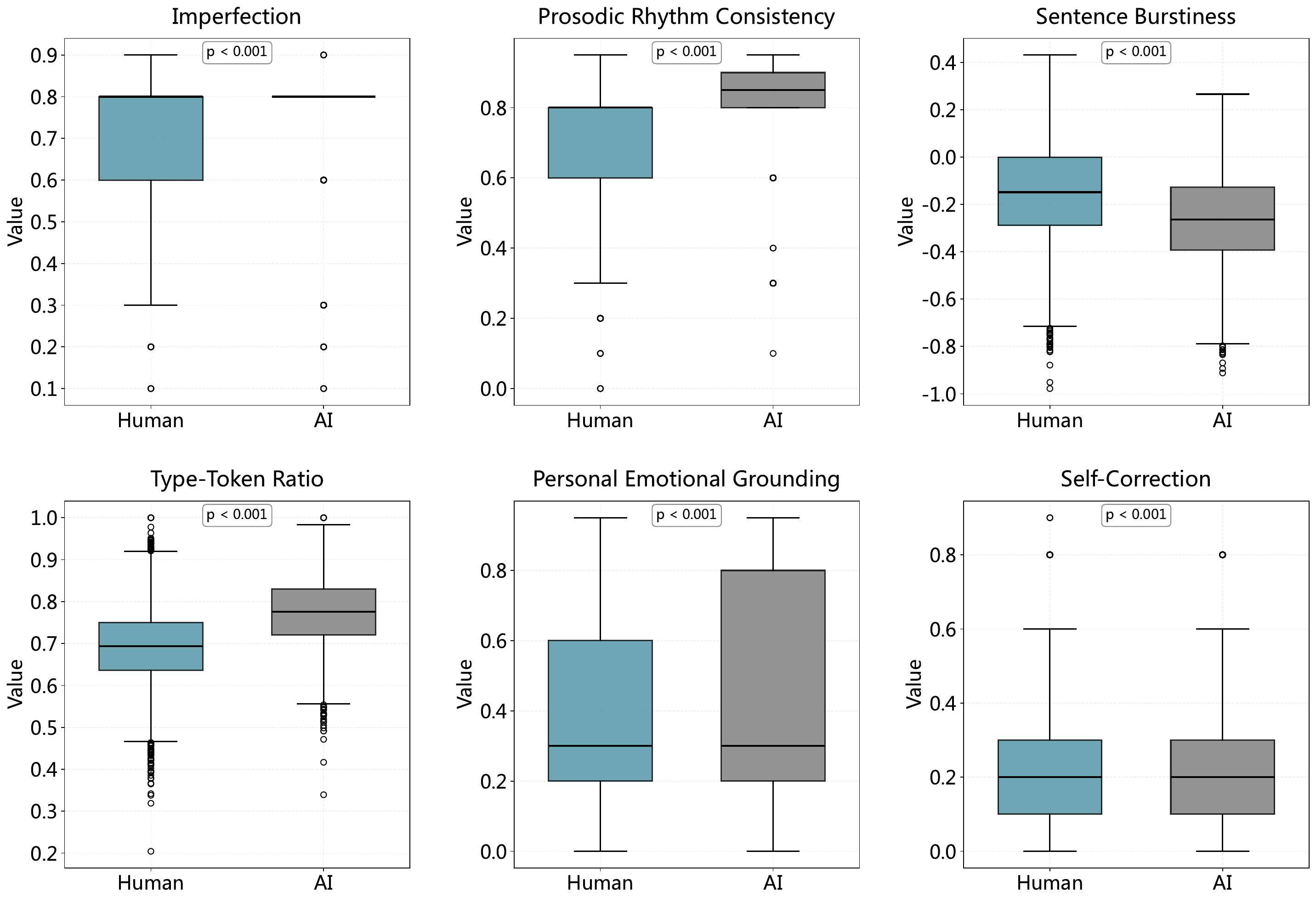}
    \caption{Comparison of feature statistics between human and AI-generated texts.}

    \label{fig:feature_stat}
\end{figure*}

To understand the differences in text style between human and AI, we conduct a comparative analysis of linguistic features extracted from human- and AI-authored texts. We apply statistical tests across the dataset, and we select six representative features for illustration, covering the four dimensions of our framework. As shown in Figure~\ref{fig:feature_stat}, these features highlight systematic divergences between human and AI writing styles.

The results indicate that AI texts consistently achieve higher values in imperfection, reflecting stable fluency and a lack of surface-level flaws. By contrast, human writing displays a much broader distribution. A similar contrast appears in prosodic rhythm consistency and sentence burstiness. AI-generated texts demonstrate regularity and uniformity in rhythm and fluctuations in sentence patterns, whereas human writing is more dynamic and irregular, often breaking rhythmic patterns. In the lexical level, the type-token ratio results show that AI text tends to maintain higher lexical diversity, reflecting stochastic generation processes that avoid frequent repetition. In contrast, human writers employ more shallow lexical distributions, which is shaped by personal stylistic habits.

In terms of emotion and rhythm, AI can achieve higher scores than human writing. This suggests that AI can mimic or even outperform humans in many human-like writing characteristics, while real human writing is more restrained.

\subsection{Feature Analysis for Engagement}

\label{sec:feat_analysis-2}

\begin{figure*}[!b]
    \centering
    \includegraphics[width=0.95\linewidth]{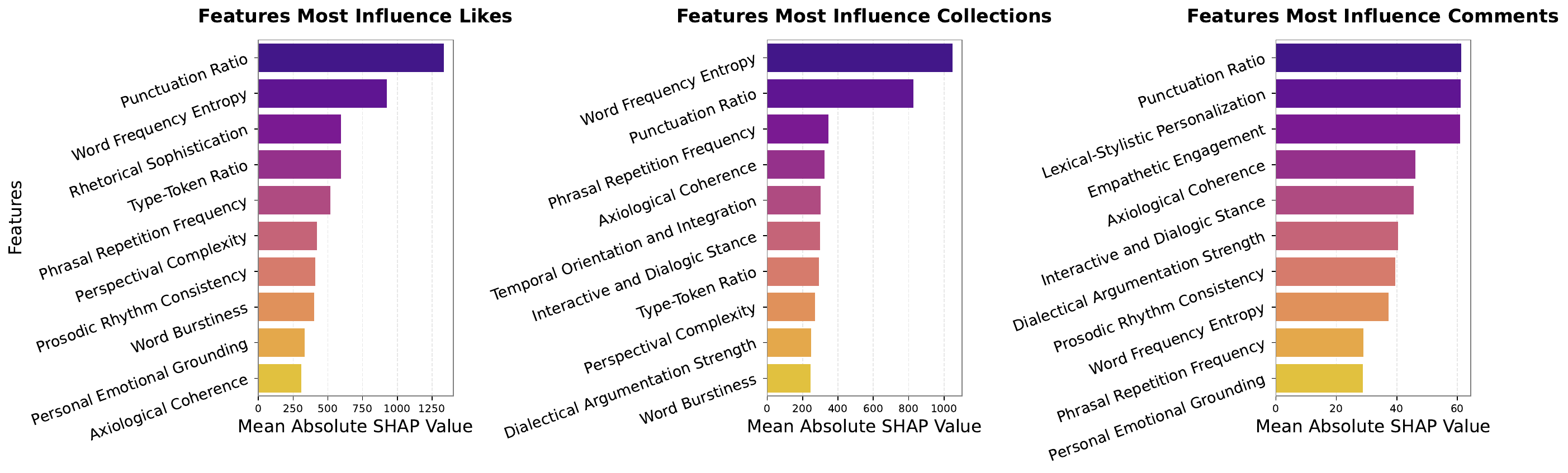}
    \caption{Top-10 most influential features of likes, collections and comments metrics.}

    \label{fig:feature_comparison}
\end{figure*}

To discover the relationship between features and engagement metrics (i.e. \texttt{likes}, \texttt{collections}, and \texttt{comments}), we conducted an analysis using SHAP value. For each metric, we train a \texttt{CatBoost} regressor model, denoted by $f(x)$, using its default hyperparameters. Subsequently, we employ SHapley Additive exPlanations (SHAP) to interpret the model's predictions \citep{lundberg2017unified}. SHAP attributes an importance value to each feature based on principles from cooperative game theory, ensuring properties like local accuracy and consistency.

As shown in Figure~\ref{fig:feature_comparison}, we observe distinct patterns of feature influence across different engagement types. For ``likes'', the most salient predictors are \textit{Punctuation Ratio} and \textit{Word Frequency Entropy}. This indicates that lightweight forms of engagement are primarily driven by surface-level features such as punctuation density and lexical diversity. In contrast, higher-order psycholinguistic features (e.g., \textit{Perspectival Complexity}, \textit{Axiological Coherence}) play only a secondary role. This suggests that likes are largely sensitive to readability and rhythm rather than deeper cognitive or semantic structures.

In the case of ``collections'', \textit{Word Frequency Entropy} emerges as the dominant feature, followed by \textit{Phrasal Repetition Frequency} and \textit{Axiological Coherence}. Compared with likes, collections are more strongly associated with content richness and value consistency. \textit{Axiological Coherence} further suggests that users are more inclined to preserve texts that demonstrate coherent values and internal logical alignment. Thus, collections appear to reflect more deliberate and evaluative forms of engagement.

For ``comments'', in addition to \textit{Punctuation Ratio}, socially oriented features such as \textit{Lexical-Stylistic Personalization}, \textit{Empathetic Engagement}, and \textit{Interactive and Dialogic Stance} exhibit the strongest influence. Unlike likes or collections, commenting behavior is primarily shaped by interpersonal dynamics, empathy, and argumentative stance. This highlights the role of dialogic and relational features in fostering deeper interactions.

The details of calculating SHAP value are shown as follow:

For a single prediction $f(x)$, SHAP explains it using an additive feature attribution model, $g(x')$:
$$
g(x') = \phi_0 + \sum_{i=1}^{M} \phi_i x'_i
$$
where $g(x') \approx f(x)$, $x'$ is a simplified binary input representing the presence ($x'_i=1$) or absence ($x'_i=0$) of a feature, $M$ is the number of features, and $\phi_i \in \mathbb{R}$ is the SHAP value for feature $i$. The term $\phi_0 = E[f(x)]$ represents the base value, which is the mean prediction over the exploring set.

The SHAP value $\phi_i$ for each feature is calculated as its marginal contribution to the prediction, averaged across all possible feature orderings (coalitions), and is formally defined as:
\begin{multline*}
\phi_i = \sum_{S \subseteq F \setminus \{i\}} \frac{|S|!(|F| - |S| - 1)!}{|F|}! \\
\times \bigl[f_{x}(S \cup \{i\}) - f_{x}(S)\bigr]
\end{multline*}

where $F$ is the full set of features, $S$ is a subset of features not including $i$, and $f_{x}(S)$ is the model's expected output conditioned on the feature values in $S$. For our analysis, we used the mean absolute SHAP value, $ \frac{1}{N}\sum_{j=1}^{N}|\phi_i^{(j)}| $, as the metric for global feature importance.

\subsection{Feature Correlation Analysis}
\label{sec:feat_analysis-3}

\begin{figure*}[!b]
    \centering
    \includegraphics[width=0.95\linewidth]{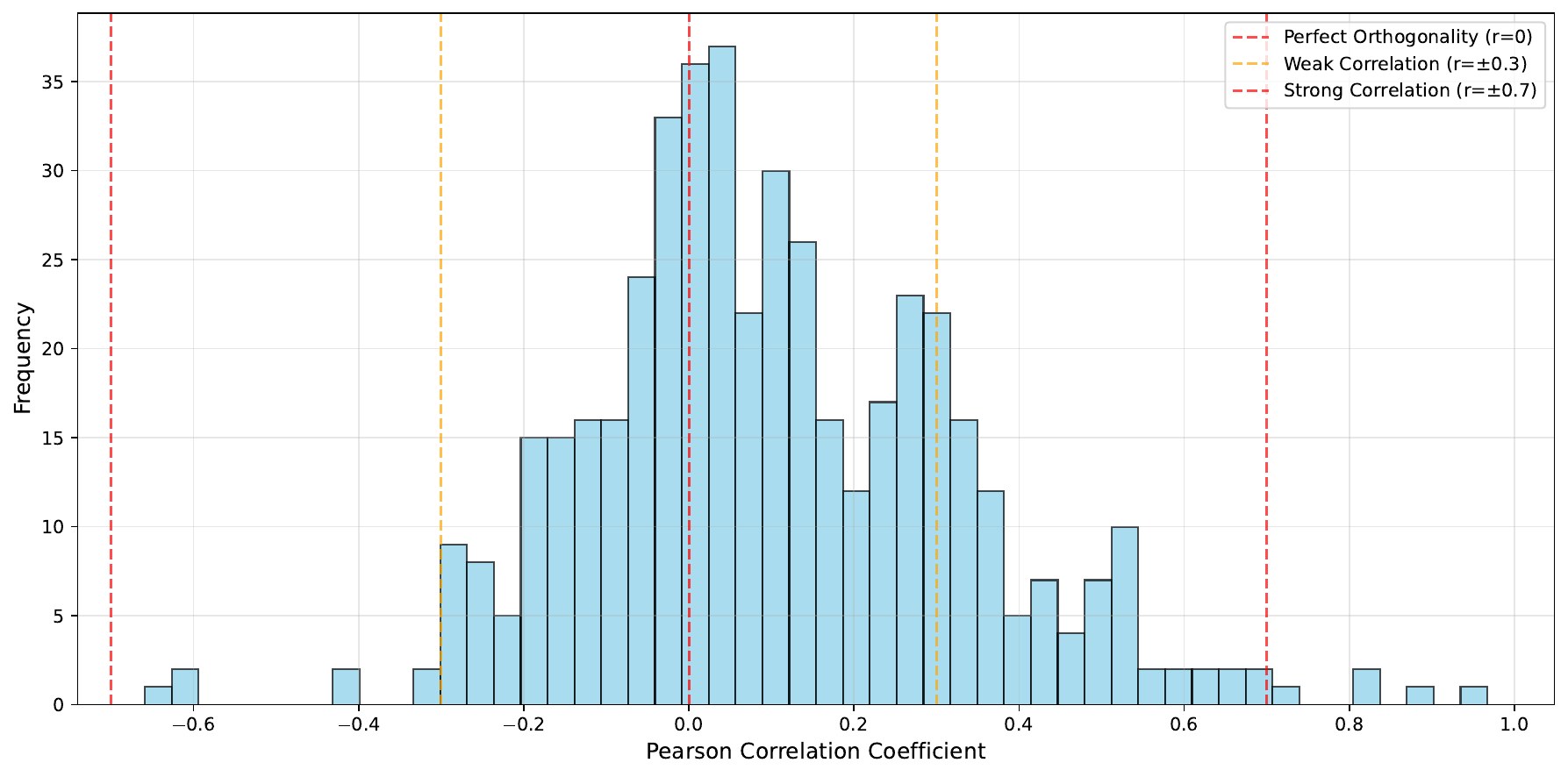}
    \caption{Distribution of pairwise feature correlations. Dashed lines indicate thresholds for weak ($|r|=0.3$) and strong ($|r|=0.7$) correlation. Most feature pairs are weakly correlated, demonstrating orthogonality of the feature set.}
    \label{fig:correlation_distribution}
\end{figure*}

To ensure that the proposed features capture complementary aspects of text, we conducted a pairwise correlation analysis. Figure~\ref{fig:correlation_distribution} illustrates the distribution of Pearson correlation coefficients across all feature pairs ($N=465$). 

Overall, the results confirm that the features are weakly correlated. The average absolute correlation is $0.1884$ with a standard deviation of $0.2290$, suggesting that the majority of features contribute orthogonal information. Specifically, only 6 feature pairs exhibit strong correlations ($|r| > 0.7$), while 89 pairs fall into the medium range ($0.3 < |r| < 0.7$). The vast majority, 370 feature pairs, remain weakly correlated ($|r| < 0.3$). 

This distribution indicates that the feature set avoids redundancy and is well-suited for capturing distinct dimensions of linguistic behavior. Therefore, the framework benefits from a diversified set of signals spanning emotional grounding, cognitive architecture, stylistic identity, and textual cohesion. It provides a robust foundation for interpretable AIGT detection.

\section{Implementation Details}
\label{sec:impl}

\subsection{AI Sample Generation Details}
\label{sec:ai_generation}

To ensure reproducibility and transparency, we provide implementation details of our AI sample generation pipeline. The generation process is controlled by a seed-based prompting strategy, where each human-authored note serves as the reference input. Given a seed note and its corresponding domain label, we construct a structured prompt (see pseudocode below) that instructs the LLM to produce a new RedNote-style post. The prompt enforces the following constraints: (1) emulate the stylistic and colloquial properties of the seed (including informal punctuation and minor grammatical imperfections common in social media), (2) maintain thematic and length consistency with the seed, while ensuring semantic novelty, and (3) output results strictly in a predefined JSON schema containing the \texttt{title} and \texttt{content} fields.

\begin{table}[h]
\centering
\small
\fbox{
\begin{minipage}{0.95\linewidth}
\textbf{Prompt Template for AI Sample Generation:}

\vspace{0.2cm}
\texttt{Reference post: \{seed\_note\}}

\vspace{0.1cm}
\texttt{Please create a new RedNote-style post based on the reference above. Requirements:}

\vspace{0.1cm}
\texttt{1. The new post should have a similar theme and topic domain (\{domain\}) but must not be identical to the reference.}

\vspace{0.1cm}
\texttt{2. Mimic the writing style of the reference, including colloquial tone, informal punctuation, and possible minor errors typical of social media.}

\vspace{0.1cm}
\texttt{3. Keep the length roughly consistent with the seed note.}

\vspace{0.1cm}
\texttt{4. Output strictly in JSON format as follows:}

\vspace{0.1cm}
\texttt{\{"title": "Post title", "content": "Post content"\}}

\vspace{0.1cm}
\texttt{Output only the JSON object, without any extra text.}
\end{minipage}
}
\caption{Pseudocode of the prompt used for AI sample generation.}
\end{table}
\noindent

We employ commercial paid APIs from multiple providers to generate the AI samples. The generation script incorporates error handling, including automated detection of extraneous markdown wrappers (i.e., \texttt{json}) and recovery via JSON string extraction. Invalid or unparsable generations are discarded to ensure dataset integrity.

Each LLM in our pool receives identical prompts and seed distributions, guaranteeing fairness across providers. Importantly, we only select seed notes published prior to November 2022 to exclude potential AI-generated contamination.

\subsection{Feature Extraction}

\label{sec:feat_extraction}

We develop an automated pipeline to extract features from each text entry using a standardized LLM-based scoring system. Each feature is defined through structured JSON criteria that specify evaluation dimensions, scoring rubrics, and key indicators. The features are listed in Table~\ref{tab:full_feature_list}. We use the latest qwen-turbo (2025-07-15) as the proxy model.

\subsubsection{Scoring Framework}

Our approach employs a two-stage process: (1) text preprocessing involving removal of extraneous characters and normalization, and (2) LLM-based evaluation using dynamically assembled prompts. Each of the 31 features is defined by a JSON schema containing:

\begin{itemize}
    \item \textbf{Dimension description}: Definition of the psychological construct.
    \item \textbf{Scoring criteria}: Anchored 0-1 scale with explicit behavioral markers.
    \item \textbf{Key indicators}: Textual evidence to focus evaluation.
    \item \textbf{Few-shot Examples}: A set of text samples paired with their expert-assigned scores. These examples guide the model's in-context learning, calibrating its judgment to align with human evaluation standards.

\end{itemize}

For each text sample, we dynamically construct evaluation prompts by embedding the target text and relevant feature criteria into a standardized template that instructs the LLM to follow a Chain-of-Thought reasoning process.

\subsubsection{Prompt Template Structure}

The evaluation prompt is dynamically constructed by assembling five core components into a standardized template. It begins by establishing the Task Context to define the psycholinguistic analysis objective, which identifies the evaluation task. The template then assigns a Role Definition, positioning the LLM as an expert evaluator. Subsequently, the specific Dimension Specification is injected from the JSON file, followed by the preprocessed Target Text for evaluation. Finally, the prompt provides detailed CoT Instruction, guiding the LLM through reasoning steps with metacognitive checks to ensure a rigorous scoring process.

The LLM is instructed to output only a numerical score between $0.0$ and $1.0$, ensuring standardized quantitative assessment across all features.

\subsubsection{Example: Emotional Intensity}

To illustrate our approach, we present the JSON example for emotional intensity evaluation:

\begin{lstlisting}[basicstyle=\ttfamily\small]
{
  "dimension_id": "emotional_intensity",
  "description": "Evaluates the depth, regulation, and contextual appropriateness of emotional expression. This dimension assesses the presence of emotion, and its nuance, variability, and alignment with the narrative events. High scores reflect a rich, well-regulated, and contextually congruent emotional landscape, while low scores indicate expressions that are flat, extreme, or mismatched with the situation.",
  "scoring_criteria": {
    "0_score": "Emotionally flat, suppressed, or chaotically unregulated. The expression is either monotonous (e.g., alexithymic, detached) or extreme and overwhelming (e.g., hysterical, disproportionate rage). There is a significant incongruence between the emotion described and the context.",
    "1_score": "Rich, nuanced, and contextually appropriate emotional expression. The author conveys a spectrum of feelings using diverse vocabulary. Emotional intensity is well-regulated, rising and falling in a way that is congruent with the narrative. Acknowledges complex or mixed emotions."
  },
  "key_indicators": [
    "Analyze the ratio and distribution of positive (e.g., 'joy', 'relief') vs. negative (e.g., 'grief', 'fear') emotion words. Assess the mix of high-arousal (e.g., 'ecstatic', 'furious') vs. low-arousal (e.g., 'serene', 'content') terms.",
    "Evaluate the richness of the emotional lexicon. Does the author use a variety of synonyms and descriptors for feelings, or repeatedly use the same basic emotion words?",
    ...
  ],
  "few-shot examples": [
    {
      "text": "I can't believe she left. I'M SO ANGRY! EVERYTHING IS AWFUL! I will NEVER be happy again, this is the worst thing that could ever happen to anyone! I hate everything and everyone!",
      "score": 0.3,
      "rationale": "While strong emotion is present, it is extreme, one-dimensional, and unregulated. The use of absolutes ('NEVER', 'EVERYTHING') and disproportionate intensity without nuance suggests a lack of emotional modulation, mapping to the lower end of the scale."
    },
    ...
  ]
}
\end{lstlisting}

This structured approach ensures consistent, objective evaluation across all 31 psycholinguistic dimensions while maintaining the flexibility to adapt criteria for different psychological constructs.

\subsection{Reliability of Proxy LLMs}
\label{sec:reliability_of_proxy_llms}
Psychometrics using LLMs is a rapidly evolving field, and existing research has demonstrated that modern LLMs can be more reliable and stable for psychological text analysis than traditional tools. In this work, we adopt the established practice for feature extraction with targeted prompt engineering. Since prompt comparison has a large search space, a more meaningful comparison is the performance of different proxy models. Therefore, in this section, we utilize a human-annotated dataset of psychological constructs to evaluate and verify the performance of our proxy model.

In Table~\ref{tab:proxy_correlation}, we report the Spearman correlation of different proxy LLMs and traditional LIWC-based methods against human labels on different psychological constructs.

\begin{table}[h]
\centering
\resizebox{\linewidth}{!}{
\begin{tabular}{l|ccc}
\toprule
\textbf{Method} & \textbf{Concreteness} & \textbf{Sentiment} & \textbf{Formality} \\
\midrule
Qwen-turbo & 0.720 & 0.785 & 0.770 \\
DeepSeek-V3 & 0.716 & 0.793 & 0.768 \\
GPT-4o & 0.746 & 0.795 & 0.788 \\
LIWC-based Method & 0.554 & 0.570 & 0.429 \\
\bottomrule
\end{tabular}}
\caption{Spearman correlation of different methods against human labels on psychological constructs.}
\label{tab:proxy_correlation}
\end{table}

We acknowledge that despite using the best possible methods to extract features, existing methods cannot achieve 100\% accuracy. However, the PLAD framework does not rely on absolute values for interpretability; instead, it distinguishes human and AI text based on differences across various feature dimensions.

\end{document}